\documentclass[11pt, a4paper, confidential, copyright, gdm]{google}

\usepackage[authoryear, sort&compress, round]{natbib}
\usepackage{amsmath, amssymb, booktabs, xspace}
\usepackage{paralist}
\usepackage{enumitem}
\usepackage{multirow}
\usepackage{algorithm}
\usepackage{booktabs}
\usepackage{makecell}
\usepackage{tabularx}
\usepackage{array}
\usepackage{adjustbox}
\newcolumntype{L}{>{\raggedright\arraybackslash}X}
\usepackage{algpseudocode}
\usepackage{listings}
\newcommand{\dataset}{\mathcal{D}}

\newcommand{\tasks}{\mathcal{T}}
\newcommand{\eprecord}{\mathbf{d}}

\newcommand{\judge}{\mathcal{J}}
\newcommand{\prompt}{\pi}
\newcommand{\expertset}{\Pi^*}
\newcommand{\playbook}{\mathcal{P}}

\uselogo{} 

\title{Learning to GUIDE the Group: Self-Learning LLM Facilitation Across Multi-Party Tasks}

\newtoks\correspondingauthor
\correspondingauthor{zliu2803@usc.edu, cjqian@google.com}

\renewcommand{\today}{2026-10-02}

\author[*,1]{Ziyi Liu}
\author[2]{Tzu-Sheng Kuo}
\author[3]{Jiale Liu}
\author[1,4]{Deqing Fu}
\author[5]{Nithum Thain}
\author[5]{Crystal Qian}

\affil[*]{Work done during an internship at Google DeepMind.}
\affil[1]{University of Southern California}
\affil[2]{Carnegie Mellon University}
\affil[3]{Pennsylvania State University}
\affil[4]{Google Research}
\affil[5]{Google DeepMind}

\usepackage{subcaption}
\usepackage{pifont}
\usepackage[T1]{fontenc}
\usepackage{xcolor}
\newcommand{\yes}{\textcolor{green!60!black}{\ding{51}}}
\newcommand{\no}{\textcolor{red!70!black}{\ding{55}}}
\usepackage{wasysym}
\begin{abstract}
As large language models (LLMs) are increasingly adopted as multi-party communication facilitators, there is a growing demand for agents capable of balancing diverse group dynamics across contexts. However, existing LLM facilitators are often fine-tuned for task-specific objectives; textit{generalized} LLM-based facilitation methods that can adapt to different types of tasks and diverse group dynamics are underdeveloped. To address this gap, we present \textsc{GUIDE} (\textbf{G}eneralized \textbf{U}nified \textbf{I}ntervention for \textbf{D}ynamic \textbf{E}xchanges), a framework for developing generalized LLM-based facilitators across different multi-party contexts, such as open-ended discussion, negotiation, and consensus-building. \textsc{GUIDE} evaluates facilitation across both objective task outcomes and subjective metrics, including psychological safety and perceived friction. Within this framework, we study whether LLMs can self-learn generalized facilitation strategies offline to real-time interventions. We first collect a corpus of multi-party human interactions guided by task-specific expert LLM facilitators (N=432 conversations). Grounded in these dialogues, we then employ a self-learning method in which an LLM induces an actionable strategic playbook and iteratively refines it using feedback from an LLM judge aligned with human annotators (N=20), developing a single generalized facilitator that outperforms the three task-specific expert facilitators on
held-out conversations ($61.4\%$ win rate). Finally, we evaluate these improvements empirically; in a real-time online empirical study (N=462 conversations), the self-learned generalized facilitator
preserves facilitator performance and group productivity, improves speaking
comfort and reduces interpersonal friction, while using $24.7\%$ fewer turns
and $29.0\%$ fewer words than the expert prompts. These findings provide empirical evidence that LLMs can learn generalizable, task-agnostic social facilitation from task-specific human interactions, on-par with hand-crafted task-specific prompts across diverse group settings.

\end{abstract}

\begin{document}

\maketitle

\section{Introduction}

As artificial intelligence systems increasingly transition from isolated one-on-one assistants to collaborative partners in human communication~\citep{10.1145/3805689.3812253, Houtti_2025}, there is a growing demand for LLM agents capable of generalizing across diverse, multi-party collaborative scenarios~\citep{alsobay2025bringing, 10.1145/3742413.3789078, chen2026simulating,liu2026teamfusion}. Unlike in one-to-one settings, multi-party facilitation agents must simultaneously navigate diverse participant interaction dynamics, competing individual and group outcomes and perceptions, and flexibly adapting across collaboration styles and objectives.

Existing research on AI-facilitated group interactions remains largely constrained to narrow, task-specific domains with an overemphasis on immediate task outcomes~\citep{liu-etal-2026-promediate, chen2026simulating}. This paradigm suffers from several shortcomings. First, domain-specific systems rely heavily on manual prompt engineering, which involves substantial trial-and-error that can fail to cover unpredictable conversational corner cases. Second, there's an absence of validated empirical data in this space; prior evaluations either involve underpowered, small-scale human studies or large-scale multi-agent simulations that fail to reflect authentic human dynamics~\citep{stephenson2025comparing, bojic2026llmagentspredictsocial, kwon-etal-2025-evaluating, kwon2026llmstrulyembodyhuman}. Finally, by treating transactional outcomes (e.g., immediate agreement) as the primary success metric, prior work overlooks relational, interpersonal signals.
\begin{figure*}[t]
\centering
\includegraphics[width=\textwidth]{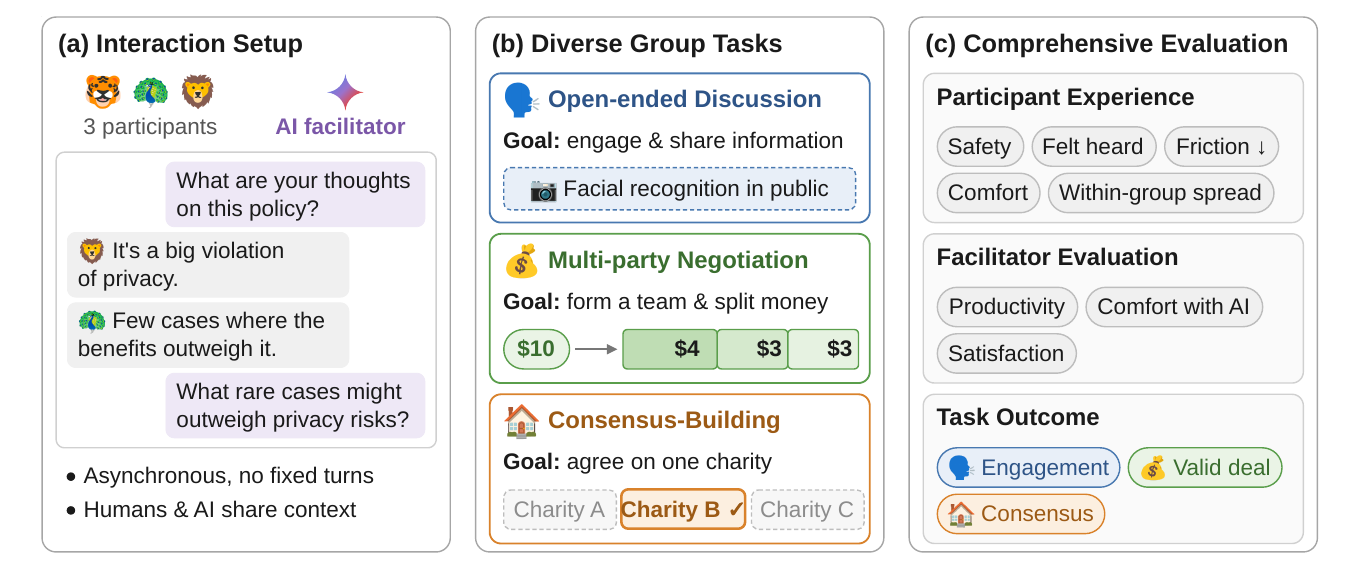}
\caption{\textbf{GUIDE overview.} \textbf{(a)} Three participants interact with an AI facilitator. \textbf{(b)} Each group completes open-ended discussion, multi-party negotiation, and consensus-building tasks in random order. \textbf{(c)} Evaluation covers participant experience, facilitator ratings, and task-specific outcomes.
}
\label{fig:framework}
\end{figure*}

In this study, we evaluate \textit{whether an LLM agent self-learn a generalized facilitation strategy capable of navigating diverse group dynamics across different types of collaborative tasks}. We introduce \textsc{\textbf{GUIDE}} (\textbf{G}eneralized \textbf{U}nified \textbf{I}ntervention for \textbf{D}ynamic \textbf{E}xchanges): a self-learning framework that develops generalized facilitation strategies, informed by a corpus of large-scale real-world human conversations guided by task-specific expert facilitators. We evaluate the learned facilitation in a real-time empirical evaluation involving three types of tasks --- open-ended discussion, multi-party negotiation, and consensus-building --- each with distinct interaction goals and outcome structures. In addition to optimizing for (1) outcome-level metrics (such as whether a deal was achieved in the negotiation setting or whether participants could agree during consensus-building), the framework evaluates (2) subjective participant experience and (3) subjective perceptions around the AI facilitator (see Table \ref{tab:comparison_with_prior} for comparison with previous work). 
Concretely, the framework produces a \emph{playbook}: an explicit, human-readable set of facilitation rules that an LLM revises on its own, generation after generation, informed by a corpus of real, facilitated multi-person discussions we collected as learning material.

Unlike conventional self-improving agents designed for domains with automated verifiers~\citep{zhang2026agentic, shinn2023reflexion, fernando2023promptbreeder,pei2026scopepromptevolutionenhancing,tao2026seposelfevolvingpromptagent}, our framework operates in an open-ended social setting where the impact of an intervention cannot be objectively validated or counterfactually simulated with sufficient fidelity~\citep{lu-etal-2026-llm, li2026aiagentsyetsufficient}. Our framework improves iteratively through version-over-version comparison rather than direct supervision against the expert: at sampled conversational turns, an LLM judge aligned to human annotators evaluates candidate interventions purely against those from the previous generation's playbook in a blind pairwise preference test. These per-turn preferences, attributed back to the individual rules the model cited, provide the learning signal. Direct pairwise comparisons against the expert facilitator are reserved strictly as an unseen benchmark on the test set (Figure \ref{fig:algo}).

Our empirical evaluations demonstrate the efficacy of the learned facilitator across both offline verification and human study. First, on held-out conversations evaluated by our aligned judge, the learned playbook outperforms all three expert baselines, achieving a 61.4\% win rate over 736 decision points across domains. Second, when deployed in $154$ groups to facilitate $462$ real-time conversations, the self-learned facilitator actively improves the participant experience compared to the expert facilitators. Groups guided by the self-learned facilitator report higher psychological safety and a greater sense of feeling heard, alongside significantly decreased friction and anger. Crucially, the self-learned facilitator significantly decreases the dispersion of these experiential metrics among human participants within the same group. This reduction in variance could indicate that the agent fosters a more equitable conversational environment, ensuring that all members share a similarly supported experience. Remarkably, the self-learned facilitator achieves these improvements in fairness and overall experience while intervening $25\%$ less often per episode and generating $29\%$ fewer words.
In summary, our main contributions are:
\begin{compactitem}
\item \textbf{Generalized facilitation definition with multi-objective optimization:} We formalize conversational facilitation across diverse tasks and evaluate systems based on participant experience, facilitator perception, and task success.
\item \textbf{Human-grounded learning framework:} We develop a self-learning framework trained on multi-party conversations with human-expert facilitation prompts, using a human-aligned LLM judge ($\kappa =0.70$) to provide turn-level self-improvement evaluation.
\item \textbf{Empirical study evaluating generalized vs. bespoke facilitation:} The self-learned facilitator improves participant experiences, reduces friction, and fosters conversational fairness.
\end{compactitem}
\begin{table}[htbp]
    \centering
    \resizebox{0.95\textwidth}{!}{%
    \begin{tabular}{lcccccc}
    \toprule
    \textbf{Method} & \textbf{Task} & \textbf{Proactive}   & \textbf{Subjective Eval.} & \textbf{> 2 Parties} & \textbf{Real Human Eval.} & \\
    \midrule
   GRAIL \cite{alsobay2025bringing} & Decision Making& \no & \LEFTcircle& \yes & \yes \\
    Observe, Ask, Intervene \cite{Houtti_2025}& Decison Making&\yes & \yes &\yes &\yes  \\
    LLMediator \cite{westermann2023llmediatorgpt4assistedonline}&Dispute Resolution & \no& \no&\no & \no\\
    ProMediate \cite{liu-etal-2026-promediate} & Negotiation&\yes &\no &\yes &\no  \\
    TeamFusion \cite{liu2026teamfusion} & Decision Making &\no &\yes &\yes &\yes  \\
    Charity Allocation \cite{10.1145/3805689.3812253} &Decision Making &\yes &\yes &\yes &\yes \\
    Interview facilitation \cite{abulimiti-etal-2026-reinforcement}&Interview&\no&\no&\no&\no \\
    Choose Your Agent \cite{zhu2026chooseagenttradeoffsadopting}&Negotiation&\no&\yes&\yes&\yes\\
    \midrule
    \textbf{GUIDE} (Ours) & Multiple Domains&\yes &\yes &\yes &\yes  \\
    \bottomrule
    \end{tabular}%
    }
    \caption{Comparison of prior LLM-based facilitation methods. \yes~ indicates full support for a criterion, \LEFTcircle~ indicates partial support, and \no ~ indicates no support.}
    \label{tab:comparison_with_prior}
\end{table}
\section{Task Formulation}
\label{sec:task_formulation}

In this section, we formulate our benchmarking framework for AI-facilitated group deliberations. We first introduce the interaction setup for LLM-facilitated multi-party conversations (\S\ref{sec:llm_facilitation}). We then present three distinct task scenarios designed to capture diverse conversational dynamics and communicative objectives (\S\ref{sec:scenarios_setup}). Finally, we detail the multi-dimensional evaluation protocol employed to assess conversational efficacy (\S\ref{sec:eval_metrics}). Figure~\ref{fig:framework} summarizes the framework end to end.

\subsection{LLM Facilitation in Multi-Party Conversation}
\label{sec:llm_facilitation}

Each multi-party interaction consists of a group of 3 human participants paired with an autonomous LLM facilitator in a asynchronous, text-based discussion environment. Participants engage in real-time without strict round-robin constraints or prescribed turn-taking order. Unlike conventional conversational AI settings where the agent holds exclusive turn priority~\citep{westermann2023llmediatorgpt4assistedonline}, acts only when explicitly summoned via direct mentions~\citep{Houtti_2025}, or speaks at fixed, predetermined intervals~\citep{liu-etal-2026-promediate}, our facilitator can proactively intervene within the unfolding dialogue (see Appendix \ref{ssec:app_flow} for implementation). 

\subsection{Scenario Configurations}
\label{sec:scenarios_setup}

Prior evaluations of automated facilitation predominantly focus on single-scenario benchmarks~\citep{10.1145/3805689.3812253, alsobay2025bringing, liu-etal-2026-promediate, zhu2026chooseagenttradeoffsadopting}, but do not evaluate an agent's ability to generalize across diverse multi-party conversational settings. To evaluate multi-domain adaptability, we design three real-time scenarios characterized by distinct collaborative objectives, constraint structures, and success criteria (full details are in Appendix \ref{sec:appendix_design}.)
\begin{compactitem}
    \item \textbf{Open-Ended Discussion.} A goal-free discussion where participants discuss policy without the need to reach an agreement; participants can receive a bonus payment for ``high-quality engagement'' (with message volume as a proxy measurement).

\item \textbf{Multi-Party Negotiation.} A negotiation game where participants must form a team and divide a pool of real money under rules that grant unequal bargaining power and prevent an even split; if participants agree on the outcome, the pool is given as a bonus payment.\footnote{Adapted from standard multi-stakeholder bargaining exercises developed by the Program on Negotiation (PON) at Harvard Law School.}. 

\item \textbf{Consensus-Building.} Participants must reach a unanimous agreement on a charity from a list of options to receive a donation; if participants agree, the donation is made.
\end{compactitem}
\subsection{Evaluation Metrics for Conversation Tasks}
\label{sec:eval_metrics}

Prior work on real-time AI intervention focuses on singular objectives such as task completion~\citep{10.1145/3742413.3789078, liu-etal-2026-promediate}. However, group dynamics introduce additional social objectives; for example, an agent can push a group to a deliverable while leaving participants talked over, or keep everyone comfortable while achieving nothing~\citep{bush1994promise, 10.1145/3805689.3812253}. We therefore evaluate each session across three practical dimensions: \emph{participant experience} (how participants perceived the interaction), \emph{facilitator evaluation} (how participants perceived the AI facilitator), and \emph{task outcome}. Subjective measures were collected immediately after each session using 5-point Likert scales alongside free-text rationales from all three participants. The specific metrics within each dimension are detailed in Table~\ref{tab:metrics} and summarized below:

\begin{compactitem}
    \item \textbf{Participant Experience:} Psychological safety, feeling heard, comfort speaking, friction/anger, and perceived group productivity.
    \item \textbf{Facilitator Evaluation:} Perceived facilitator helpfulness, overall satisfaction with the facilitator, and comfort interacting with the facilitator.
    \item \textbf{Task Outcome:} Unanimous agreement in consensus-building task, valid team allocation for the negotiation task, and  participant message volume for the open-ended conversation.
\end{compactitem}

\section{Learning a Generalized Facilitation Prompt}
\label{sec:method}
\begin{figure}
    \centering
    \includegraphics[width=0.96\linewidth]{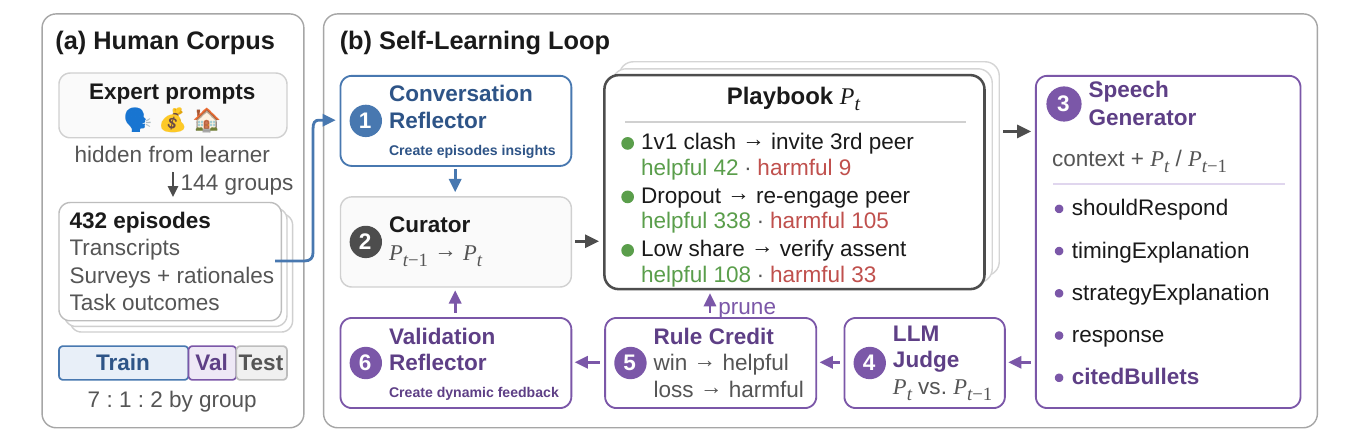}
    \caption{\textbf{Self-learning framework.} \textbf{(a)} Real human conversations collected with the expert prompts forms the training corpus. \textbf{(b)} A conversation reflector extracts lessons from human experience; a curator edits the playbook; an aligned LLM judge compares the candidate with the preceding playbook; rule-level credit attributes each cited rule's contribution; and a validation reflector converts validation failures into new lessons for the next cycle.}
    \label{fig:algo}
\end{figure}
Prior self-evolving approaches typically rely on an explicit reward, verifier, or other deterministic signal to determine whether a policy update is beneficial~\citep{zhang2026agentic, liu2023trainingsociallyalignedlanguage, kong2026improvingbehavioralalignmentllm}. Facilitation offers no such signal: whether and when an intervention is helpful depends on human perceptions, conversational dynamics, and downstream group outcomes. We therefore adapt context-evolution approaches to this setting by grounding self-improvement in a corpus of real human interactions. Our framework first reflects on these interactions to discover facilitation insights, then iteratively refines them using an LLM judge aligned with human preferences.

\subsection{Stage 1: Corpus Collection}
\label{ssec:study1}

To construct the learning corpus and establish a strong baseline, we manually designed the expert prompts, grounding them in prior work~\citep{bush1994promise, govers2024ai, liu-etal-2026-promediate, 10.1145/3805689.3812253} and refining them through pilot studies with real participants. Using these prompts, we recruited $432$ participants forming $144$ non-overlapping teams of three. Each team completed all three tasks in Section~\ref{sec:scenarios_setup}, yielding $432$ interaction episodes. For each episode $e$, we record the task context $c_e$, conversation transcript $x_e$, participant feedback $q_e$, and task outcomes $o_e$. We split the data at the team level into training, validation, and test sets using a $7{:}1{:}2$ ratio. Full study procedures and dataset statistics are provided in Appendix~\ref{ssec:app_corpus}.

\subsection{Self-Learning Framework}
\label{ssec:policy}

\paragraph{Playbook-based policy.}
We keep the underlying language model frozen and represent its facilitation strategy using a structured natural-language \emph{playbook} $\playbook$. We initialize from an empty playbook, $\playbook_0=\varnothing$, with only a minimal instruction describing the facilitator's role and requiring it to decide both \emph{whether} to intervene and \emph{what} to say.

At each participant turn, the facilitator observes the interaction history $h_i=(c_e,x_e^{\le i})$ and produces $a_i\sim\pi_{\playbook}(h_i)$, where $a_i=(\rho_i,\delta_i,\mu_i,C_i)$, $\rho_i$ is the model's decision rationale, comprising its timing and strategy explanations; $\delta_i\in\{\textsc{Speak},\textsc{Silent}\}$ is the intervention decision; $\mu_i$ is the generated utterance and is empty when $\delta_i=\textsc{Silent}$; and $C_i\subseteq\playbook$ contains the playbook rules cited to support the decision. Explicitly recording $C_i$ enables us to attribute validation feedback to individual rules.

\paragraph{Human-experience reflection.}
As illustrated in Figure~\ref{fig:algo}, learning begins with a \emph{conversation reflector}. At generation $t$, it reads a batch of episodes containing conversation transcripts, participant feedback, and task outcomes, then extracts transferable lessons about effective and ineffective facilitation. This feedback path grounds learning in real human experience. The reflector does not observe the expert prompts or their rationales as demonstrations.

\paragraph{Curator.}
The \emph{curator} converts the reflector's lessons into local edits of the incumbent playbook $\playbook_{t-1}$, producing a candidate $\widetilde{\playbook}_t$. It may add, edit, merge, or remove individual rules rather than rewrite the entire playbook. From the second generation onward, the curator also receives lessons from the validation reflector described below, combining static evidence from the human corpus with dynamic feedback about the current playbook.

\paragraph{LLM judge.}
Because facilitation has no deterministic reward, an LLM judge $\judge$ provides the validation signal. We manually calibrate its rubric against human annotations before learning, reaching Cohen's $\kappa=\mathbf{0.70}$ agreement with the human panel (details in Appendix \ref{sec:appendix_rubrics}). At each held-out validation state $h_i$, the speech generator produces one action from the candidate playbook and one from the preceding playbook, $a_i^{t}\sim\pi_{\widetilde{\playbook}_t}(h_i)$ and $a_i^{t-1}\sim\pi_{\playbook_{t-1}}(h_i)$, and we ask the judge to perform a blind pairwise comparison: $(v_i,r_i) = \judge(h_i,a_i^{t},a_i^{t-1})$, where $v_i$ indicates whether the candidate, the preceding playbook, or neither is preferred, and $r_i$ explains the judgment. The comparison is blind and uses the same conversation state for both actions. \textbf{Crucially, neither action comes from an expert prompt:} during learning, the judge measures improvement over $\playbook_{t-1}$ rather than similarity to the expert baseline.

\paragraph{Validation reflection.}
The \emph{validation reflector} supplies the second, dynamic feedback path. It examines validation losses and the judge's rationales to identify recurring problems in the current playbook, such as unnecessary intervention, missed opportunities, poor timing, or conflicts between rules. This component is necessary because human-experience reflection is limited to a static corpus: it cannot reveal how newly added or modified rules behave when executed. By contrast, the LLM judge generates new evidence at every generation, and the validation reflector converts that evidence into lessons that the curator can use in the next cycle.

\paragraph{Rule-level refinement.}
The judge's verdict evaluates an entire action, although multiple cited rules $C_i$ may have contributed to it. We therefore use the judgment rationale to label each cited rule as \textsc{Helpful}, \textsc{Harmful}, or \textsc{Neutral}. These credits accumulate across validation examples, allowing the framework to preserve useful rules while pruning rules that are consistently harmful or unused. Rule-level attribution also prevents one poor action from indiscriminately penalizing every rule it cited.

The framework thus alternates between static human-experience evidence and dynamic policy-validation evidence. After $G$ generations, the final playbook $\playbook_G$ defines the self-learned prompt. Only then is it compared with the expert prompts on the held-out test set. Full prompts, pruning criteria, decision-point sampling, and optimization details are provided in Appendix~\ref{ssec:app_loop}.

\section{Evaluation and Results}
\label{sec:evaluation}
\label{sec:results}

We ask two research questions to formulate our experiments: 
\begin{compactitem}
\item \textbf{RQ1:} Can automated self-improvement discover a generalized facilitation strategy capable of outperforming domain-tailored prompts?
\item \textbf{RQ2:} How effectively does this learned strategy preserve deliberative quality and participant satisfaction in real-world discussions?
\end{compactitem}

\subsection{Experimental Setup}
\label{ssec:offline}
\label{ssec:study2}

\paragraph{Self-learning configuration.}
As described in Sec~\ref{sec:method}, our self-learning framework comprises five distinct components: a conversation reflector, a validation reflector, 
a curator, an LLM judge , and a speech generator which generate facilitator action given the context and the playbook.

To isolate the effects of the reflection and curation stages, we hold the remaining components constant: the LLM judge is fixed to \texttt{gemini-3.7-flash} for consistent scoring, and the speech generator is fixed to \texttt{gemini-3-flash} to match the model used during Stage~1 data collection. For the reflectors and curator, we evaluate three model backbones: \texttt{gemini-3-flash}, \texttt{gemini-3.7-flash}, and \texttt{gemini-3-pro}. We explore two temperature settings ($0.2$ and $0.8$) and three batch sizes ($B \in \{5, 10, 20\}$). The batch size $B$ dictates both the number of training conversations ingested by the reflector per step and the subsample size evaluated during validation. Because validating over the complete validation set at every step would incur prohibitive runtime (4--5 hours per validation round), we evaluate on a fixed budget of $B$ conversations per step, sampling without replacement across steps so that all validation instances are evaluated with uniform probability. Finally, we ablate the validation reflector (with vs.\ without). All configurations are trained for exactly one epoch over the Stage~1 corpus to ensure full data coverage while preventing overfitting to individual participant styles.

\paragraph{Evaluation protocol.}
Self-learning evaluation uses the held-out Stage~1 corpus split, comprising 87 episodes from 29 unseen cohorts and 736 balanced speech/silence decision points. At each point, the fixed LLM judge blindly compares the action selected by the learned policy against the logged action of the task-specific expert prompt. We report the tie-adjusted win rate against the baseline, where ties are awarded half credit and 50\% represents parity. For live evaluation, we deploy the learned prompt to 154 new three-person cohorts (462 episodes), in the same tasks, platform, post-task survey instruments, and base facilitator backbone (\texttt{gemini-3-flash}). We then perform a between-study comparison between the outcomes in Stage~1, with distinct task-specific expert LLM-facilitator prompts, and the current stage, with one generalized facilitation prompt across the three different tasks. Full design controls are provided in Appendix~\ref{sec:appendix_design}.

\subsection{Self-learning generalized facilitator vs. expert facilitator}
\label{ssec:offline_results}

The top-performing learned generalized facilitation architecture uses \texttt{gemini-3-flash} (temperature 0.8, batch size $B=10$, and both reflectors), reaching an overall win rate of $W_{\text{test}} = 61.4\%$ over the expert baselines on held-out decisions (Table~\ref{tab:offline}).\footnote{Full ablation results are provided in Appendix~\ref{sec:appendix_ablation}.} This lead is highest on turns where the baseline chose to speak ($66.9\%$), confirming that the playbook’s edge reflects high-quality interventions rather than winning by choosing not to speak.

\begin{table}[t]
\centering
\small
\setlength{\tabcolsep}{5pt}
\begin{tabular}{@{}lrr@{}}
\toprule
\textbf{Comparison between Expert and Learned} & \textbf{Value} & \textbf{$N$} \\
\midrule
Overall learned prompt vs.\ Study~1 experts & $\mathbf{61.4\%}$ & $736$ \\
Where Expert spoke & $\mathbf{66.9\%}$ & $368$ \\
Where Expert was silent & $56.0\%$ & $368$ \\
\midrule
\textbf{Prompt correspondence and contribution} & \textbf{Value} & \textbf{Support} \\
\midrule
Expert skills represented in the learned prompt & $47/97$ ($48.5\%$) & semantic alignment \\
Learned rules without an expert antecedent & $23/53$ ($43.4\%$) & novel rules \\
Active turns citing only expert-derived rules ($W_{\text{test}}$) & $52.9\%$ & $75$ \\
Active turns citing $\geq 1$ novel rule ($W_{\text{test}}$) & $\mathbf{77.8\%}$ & $177$ \\
\bottomrule
\end{tabular}
\caption{Self-learning performance and correspondence between the learned and expert prompts. $W_{\text{test}}=50\%$ is parity. The final two rows consider active, non-silence turns and exclude ties.}
\label{tab:offline}
\end{table}

\begin{figure}[t]
\centering
\begin{subfigure}[t]{0.53\linewidth}
    \vspace{0pt}
    \centering
    \includegraphics[width=\linewidth]{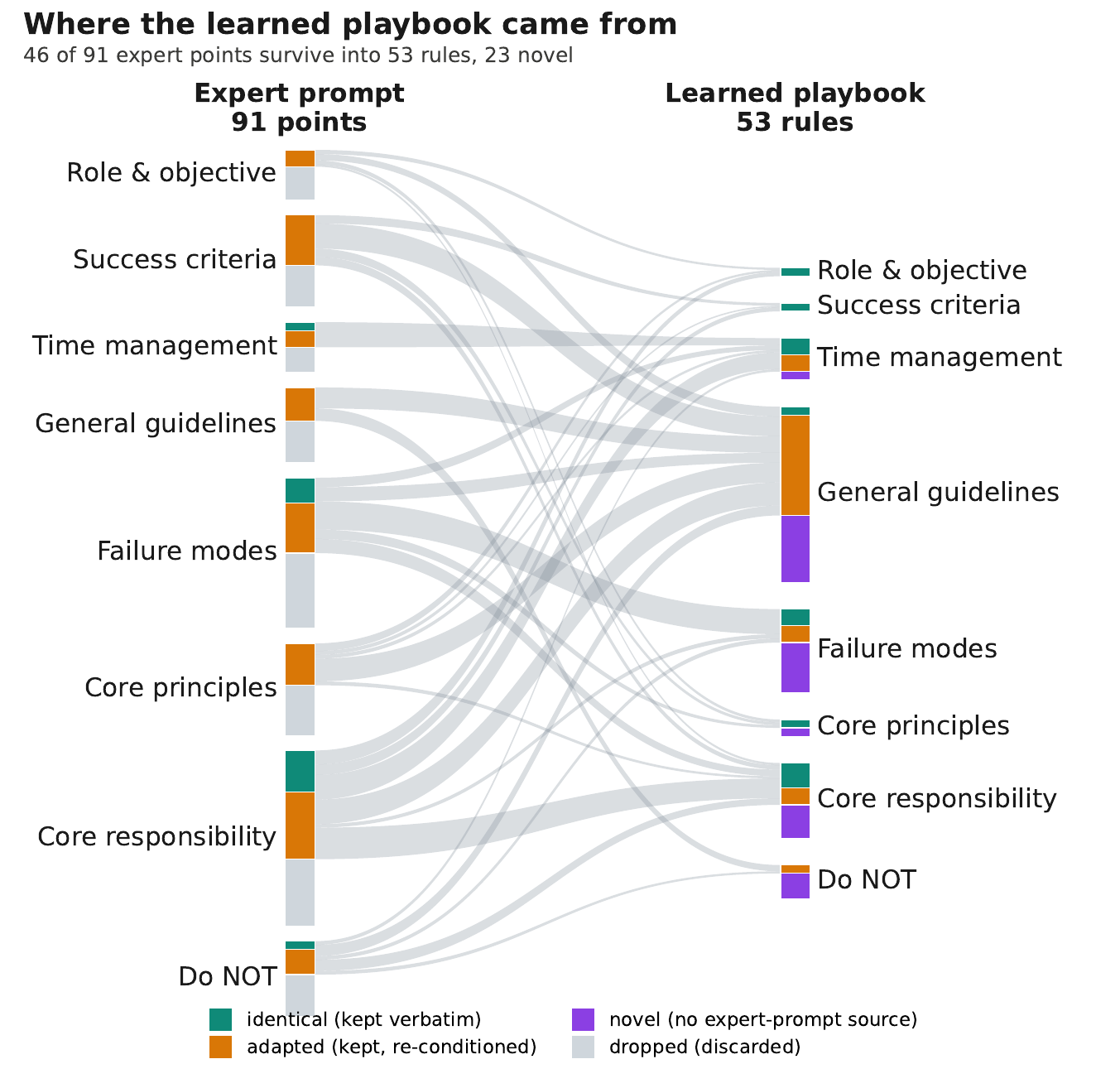}
    \caption{Expert-instruction coverage in the learned playbook.}
    \label{fig:skill_network_provenance}
\end{subfigure}\hfill
\begin{subfigure}[t]{0.45\linewidth}
    \vspace{0pt}
    \centering
    \includegraphics[width=\linewidth]{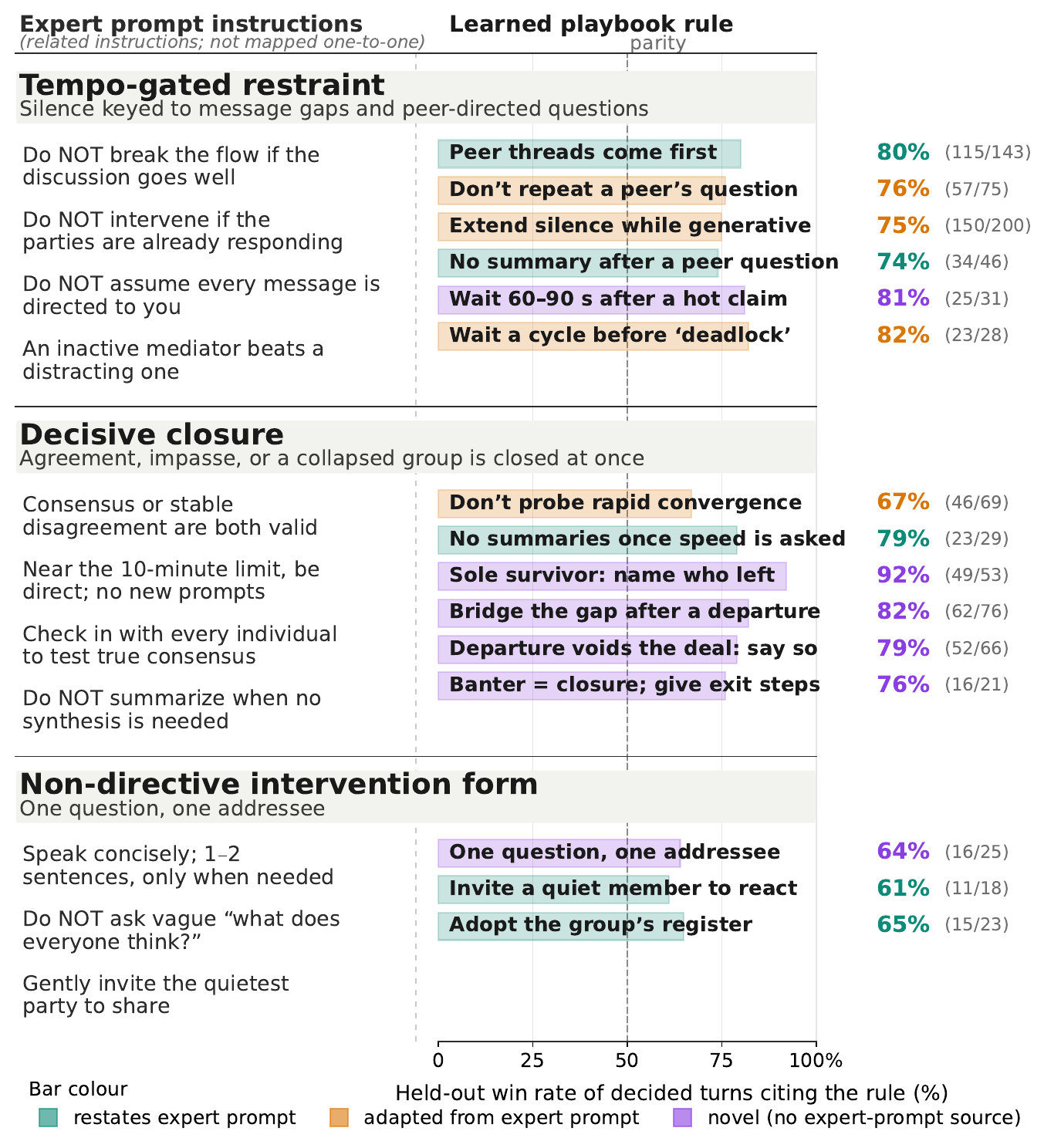}
    \caption{Expert instructions, learned rules, and held-out win rates across three behavioral dimensions.}
    \label{fig:skill_network_rules}
\end{subfigure}
\caption{Correspondence between the experts and learned playbook. \textbf{Left:} skill-to-rule alignment following the same scaffolding. \textbf{Right:} influential learned rules (simplified) span retained facilitation hygiene, state-gated intervention, and failure-mode handling.}
\label{fig:skill_network}
\end{figure}

\paragraph{Playbook learnings and additions.}
Figures~\ref{fig:skill_network_provenance} and~\ref{fig:skill_network_rules} show that, of the $91$ behavioral instructions in the expert prompts (3 expert prompts combined), $47$ survive into the $53$ learned rules, but only $10$ survive verbatim; the rest are re-conditioned, and $23$ rules have no expert-prompt source. Most improvements come not from introducing new facilitation strategies, but from specifying more precisely when existing strategies should be applied, along three dimensions. First, \emph{tempo-gated restraint} operationalizes the expert's vague instruction not to disrupt productive flow: the playbook remains silent during rapid exchanges, defers after peer-directed questions, and waits longer after controversial claims. These rules win $74$--$84\%$ of cited turns and account for $134$ of the playbook's $274$ wins, largely when the expert spoke but the judge preferred silence. Second, \emph{decisive closure} avoids re-checking or recapping agreements the group has already made and adds a capability absent from the expert prompts: handling participant departures by identifying who left, invalidating agreements that are no longer feasible, and giving the remaining members a path to closure. The departure rules win $63$ of $70$ decided turns and achieve the playbook's highest per-rule win rates ($79$--$92\%$). Third, \emph{non-directive intervention form} favors one focused question to one addressee, targeted invitations to quieter members, and questions rather than numerical templates that could anchor the group's decision; these rules win $61$--$65\%$ of cited turns. The trade-off is that the playbook sometimes accepts rapid consensus that the judge would still probe and omits a few explicit constraints, such as the ten-minute clock and integer-split rule. Overall, self-learning converts broad standing instructions into state-conditioned rules and adds robust failure-mode handling, while occasionally losing useful unconditional safeguards. More details are in Appendix \ref{ssec:app_provenance}. 


\subsection{Live Deployment: Comparable Group Experience and Facilitator Evaluation}
\label{ssec:live_results}
\begin{table}[t]
\centering
\scriptsize
\setlength{\tabcolsep}{3pt}
\scalebox{0.95}{%
\begin{tabular}{@{}lrrrrr@{}}
\toprule
\textbf{Measure} & \textbf{Expert} & \textbf{Learned} & $\boldsymbol{\Delta}$ & $\boldsymbol{p}$ & \makecell{\textbf{Within-cohort variance}\\\textbf{(S1 $\to$ S2)}} \\
\midrule
\multicolumn{6}{@{}l}{\emph{Participant experience (1--5; lower dispersion = more equitable within-group experience)}} \\
Psychological safety & $4.62$ & $\mathbf{4.67}$ & $+0.05$ & $.057$ & $0.458 \to \mathbf{0.423}$ \\
Felt heard & $4.35$ & $4.40$ & $+0.05$ & $.202$ & $0.843 \to \mathbf{0.829}$ \\
Comfortable speaking & $4.63$ & $\mathbf{4.69}$ & $+0.05$ & $.049$ & $0.430 \to \mathbf{0.407}$ \\
Friction and anger $\downarrow$ & $1.60$ & $\mathbf{1.42}$ & $\mathbf{-0.18}$ & $<.001$ & $0.812 \to \mathbf{0.663}$ \\
Sensitivity $\downarrow$ & $1.54$ & $1.46$ & $-0.08$ & $.052$ & $0.954 \to \mathbf{0.811}$ \\
Group productive & $4.09$ & $4.07$ & $-0.01$ & $.749$ & $1.081 \to 1.161$ \\
\midrule
\multicolumn{6}{@{}l}{\emph{Facilitator evaluation and prior calibration (1--5)}} \\
Pre-study AI facilitation attitude & $4.10$ & $3.93$ & $-0.17$ & $<.001$ & $0.483 \to 0.646$ \\
Calibrated productivity (post productivity $-$ pre expected productivity) & $-0.04$ & $-0.12$ & $-0.09$ & $.080$ & $1.225 \to 1.448$ \\
Calibrated satisfaction (post satisfaction $-$ pre willingness) & $+0.09$ & $+0.08$ & $-0.01$ & $.898$ & $0.976 \to 1.220$ \\
Calibrated comfort (post comfort $-$ pre expected comfort) & $+0.22$ & $+0.28$ & $+0.06$ & $.086$ & $0.654 \to 0.798$ \\
\midrule
\multicolumn{6}{@{}l}{\emph{Task outcomes}} \\
Unanimous consensus (charity allocation) & $88.2\%$ & $83.8\%$ & $-4.4$  & $.27$ & --- \\
Coalition agreed (negotiation) & $90.3\%$ & $89.6\%$ & $-0.7$  & $.85$ & --- \\
Valid allocation (negotiation) $\mid$ agreed & $84.6\%$ & $76.8\%$ & $-7.8$  & $.11$ & --- \\
Message volumn per participant& $6.14$ & $5.86$ & $-0.28$  & $.220$ & --- \\
\midrule
\multicolumn{6}{@{}l}{\emph{Facilitator footprint per episode}} \\
Facilitator turns $\downarrow$ & $5.31$ & $\mathbf{4.00}$ & $\mathbf{-24.7\%}$ & $<.001$ & --- \\
Facilitator words $\downarrow$ & $202.7$ & $\mathbf{143.8}$ & $\mathbf{-29.0\%}$ & $<.001$ & --- \\
\bottomrule
\end{tabular}
}
\caption{Live evaluation: expert facilitator ($432$ episodes) versus the learned facilitator($462$ episodes). Within-cohort variance measures disagreement among members of the same discussion. Lower variance on the participant-experience items indicates more equitable group experience. Full task-level, robustness, and attrition analyses appear in Appendix~\ref{sec:appendix_results}.}
\label{tab:live}
\label{tab:outcomes}
\end{table}

Table~\ref{tab:live} summarizes the live comparison. The learned facilitator matches the expert on feeling heard, group productivity, and every objective task outcome after multiple-testing correction. It improves comfortable speaking and substantially reduces reported friction; psychological safety is also higher ($4.67$ vs.\ $4.62$). Importantly, the learned prompt reduces within-group dispersion on all five participant-experience measures: members of the same group report more similar levels of safety, inclusion, comfort, sensitivity, and friction. These benefits use $24.7\%$ fewer facilitator turns and $29.0\%$ fewer facilitator words.

\paragraph{Calibrating facilitator evaluation.}
Before any task, we elicited each participant's initial attitude toward AI facilitation with three questions: (1) whether they believe an AI facilitator would make the discussion more productive, (2) whether they would feel comfortable with an AI in the conversation, and (3) whether they would be willing to have one. These priors are not outcomes of the intervention; they capture pre-existing expectations that can shape a later rating. In the second study, participants began with lower expectations overall ($3.93$ vs.\ $4.10$, $p<.001$). We therefore report within-person calibrated scores: post-task productivity minus expected productivity, post-task comfort minus expected comfort, and post-task satisfaction minus willingness to have an AI facilitator. The calibrated values answer whether the facilitator exceeded or fell short of each participant's own pre-task expectation; they do not turn the between-study comparison into a randomized causal estimate. We report the raw rating in Appendix~\ref{ssec:app_polarization}.


\subsection{Qualitative analysis of participant experience improvements}
\label{ssec:transfer}

We qualitatively analyzed participants’ free-text rationales to understand how perceptions shifted across conditions. Fixed procedural scripts in the expert baseline were frequently experienced as heavy-handed, coercive, and socially uncalibrated: participants noted that the facilitator felt rushed and pushy, struggled to "read the room," and failed to detect delicate social dynamics. A common grievance centered on perceived bias and selective engagement, with multiple participants reporting uneven attention—for instance, noting that "it could have targeted all participants for follow up instead of just one or two." By contrast, participants interacting with the learned facilitator emphasized its patience, perceived neutrality, and willingness to leave space for organic group discussion. Overall, 3.1\% more participants in the learned condition had zero complaints about the facilitator compared to the task-specific expert baseline.

These participant accounts directly align with the emergent rules governing the learned policy. Rather than enforcing rigid periodic interventions, artificially pushing for closure, or re-opening organic consensus, the learned prompt intervenes conditionally—acting primarily when it detects exclusion, confusion, or stalled discussion. Importantly, participants noted that simply reducing intervention frequency did not automatically elevate raw satisfaction; instead, it reframed their evaluation of facilitator productiveness. Because groups often felt capable of solving tasks independently once given the floor, the learned facilitator's restraint avoided unnecessary friction while stepping in selectively to protect quieter voices and provide targeted procedural scaffolding. This state-sensitive style demonstrates how a facilitator can use less language while markedly improving the perceived fairness and social climate of deliberation.
\section{Related Work}
\label{sec:related_work}

\subsection{AI assisting group conversation}
\label{ssec:rw_group}

Facilitation balances competing orientations rather than following a fixed procedure: practitioners alternate between evaluative and facilitative stances and between narrow and broad framings~\citep{riskin1996understanding, riskin2003decisionmaking}, while traditions diverge on core objectives---empowerment and recognition~\citep{bush1994promise}, narrative reframing~\citep{billikopf2002narrative}, or interest-based process management~\citep{article}. This contested nature motivates our multi-dimensional evaluation and our expectation of low inter-annotator agreement on specific interventions. Recent studies of LLM facilitators in live deliberation reveal a recurring perception--measurement gap: facilitation rarely improves consensus yet remains widely preferred, quietly steering outcomes and creating an ``illusion of inclusion'' without increasing participation equity~\citep{10.1145/3805689.3812253, alsobay2025bringing}. Related efforts target inclusive meetings~\citep{Houtti_2025}, dispute resolution~\citep{westermann2023llmediatorgpt4assistedonline}, and human--AI bargaining~\citep{10.1145/3742413.3789078}, aided by shared platforms~\citep{qian2025deliberatelabplatformrealtime}; however, most remain limited to single scenarios with hand-crafted prompts. When live trials are infeasible, evaluations often substitute simulated interlocutors~\citep{liu-etal-2026-promediate}. Yet LLM groups diverge from human group dynamics~\citep{chuang2024wisdompartisancrowdscomparing, stephenson2025comparing}, persona agents struggle to beat simple classifiers at predicting social reactions~\citep{bojic2026llmagentspredictsocial}, economic behavior replicates only partially~\citep{repec:spr:grdene:v:34:y:2025:i:6:d:10.1007_s10726-025-09946-9}, and agents systematically miss conflict dynamics~\citep{kwon-etal-2025-evaluating, kwon2026llmstrulyembodyhuman}. Behavioral-alignment methods~\citep{wang2025implicitbehavioralalignmentlanguage, kong2026improvingbehavioralalignmentllm, wu2026humanlmsimulatingusersstate} narrow but cannot close these gaps, directly degrading facilitation judgments. We therefore learn from logged human deliberation and validate through a live user study.
\subsection{AI Facilitation Improvement}
\label{ssec:rw_improve}

Most deployed facilitators rely on hand-authored prompts tuned for a single scenario, an approach that scales poorly and misses conversational corner cases. Automatic context optimization offers an alternative: Agentic Context Engineering~\citep{zhang2026agentic} evolves a structured context through reflection and incremental curation. We adopt this decomposition, but existing self-improvement methods generally assume a verifiable objective; even bootstrapping from expert demonstrations~\citep{silver2016mastering} requires a reliable evaluator that facilitation lacks. Although simulation can manufacture such a signal~\citep{liu2023trainingsociallyalignedlanguage}, the sim-to-real limitations above motivate grounding our framework in real three-person deliberations and post-episode human judgments. We use an LLM judge only as a per-decision proxy, calibrating it against human annotators before learning ($\kappa$: $0.70$) and attributing feedback to individual rules. Our approach therefore differs in generality (one policy across tasks), grounding (real rather than simulated interactions), and validation (offline and live), revealing that a policy can win offline while still receiving weaker facilitator ratings from participants (\S\ref{ssec:transfer}).
\section{Conclusion}
\label{sec:conclusion}


We show that a frozen LLM can self-learn a generalizable facilitation policy from a corpus of real human interactions, without access to the expert prompts that generated them or any deterministic reward signal. The resulting playbook — a structured natural-language policy of 53 rules induced from an empty initialization — outperforms three domain-specific expert prompts on held-out decisions (61.4\% win rate) without collapsing onto any single task type. Deployed to 462 live episodes with new participants, the learned policy preserves all task outcomes, improves speaking comfort and reduces interpersonal friction, while using 25\% fewer turns and 29\% fewer words. 

\section{Limitation}
A limitation of our offline optimization is that it evaluates decisions independently, rewarding restraint at each turn without modeling the cumulative effect of sustained silence on participants' experience of the facilitator. We recommend that offline evaluations of interactive agents report the action-rate distribution they induce alongside pairwise win rates. More broadly, the reflector-curator-validator loop is not specific to facilitation: it offers a general recipe for inducing structured policies in domains where success depends on human perception and no programmatic reward exists.

\section{Ethics Statement}

Both live studies were approved through an institutional IRB process. We recruited 894 adults through Prolific (432
in Study~1 and 462 in Study~2; no one took part in both) with English fluency as the sole inclusion criteria. All gave informed
consent before any task and could leave at any time (consent text in
Appendix~\ref{ssec:app_recruit}). Each received \$15 for a session of about one
hour plus the bonuses announced in advance. Participants were told before any task that
an AI facilitator would join. The facilitator was never presented as human, and
a debriefing disclosed the purpose of the research.

We consider the study minimal risk: discussions were short, text-only, and held
under pseudonymous animal aliases. We collected no demographics beyond an
occupational category, and no model ever received Prolific identifiers. Because the open-ended motions are politically contested, the
facilitator was instructed to remain neutral. Participants' stances moved only
slightly toward the motions ($+0.15$ on a 1--5 scale in both studies).
Facilitation agents could be misused to steer group decisions. Our learned
policy is a human-readable rulebook, and every decision cites the rules it
applied, so its behavior can be audited. It is nonetheless a research
prototype: it has been tested only on English-language tasks with Prolific
participants and has not been validated for high-stakes settings.

\bibliography{main}
\appendix

\section{Study Design, Materials, and Participants}
\label{sec:appendix_design}

This appendix documents the experimental apparatus shared by Study~1 and Study~2:
recruitment and compensation (\S\ref{ssec:app_recruit}), the stage-by-stage
participant flow (\S\ref{ssec:app_flow}), the three task materials
(\S\ref{ssec:app_tasks}), the survey instruments (\S\ref{ssec:app_surveys}), the
realised sample and its accounting (\S\ref{ssec:app_sample}), and participant
background (\S\ref{ssec:app_demog}). Every figure quoted here is recomputed from
the released data rather than from the study plan; \S\ref{ssec:app_discrepancies}
states the denominators used throughout.

Both studies ran on \emph{Deliberate Lab}~\citep{qian2025deliberatelabplatformrealtime},
an open-source platform for synchronous multi-party experiments. The two studies used identical stage
configurations; the \emph{only} manipulated element is the body of the
facilitator's system prompt during the three conversational task stages (\S\ref{ssec:study2}).

\subsection{Recruitment, Consent, and Compensation}
\label{ssec:app_recruit}

Participants were recruited through \textbf{Prolific} and completed the study in
real time in cohorts of three. Consent was obtained on a terms-of-service stage
that gated entry. The consent text read, verbatim:

\begin{quote}\small
Thank you for your interest in this research. If you choose to participate, you
will participate in three tasks in real-time with two other participants. These
tasks include:

\smallskip
\noindent\hangindent=1.5em\hbox to 1.5em{\hss$\bullet$\hskip 0.5em}An open-discussion about policies, where high-quality participation here may result in a bonus of up to \$2,

\noindent\hangindent=1.5em\hbox to 1.5em{\hss$\bullet$\hskip 0.5em}A negotiation task, where the three of you will determine how to split a bonus of up to \$10, and

\noindent\hangindent=1.5em\hbox to 1.5em{\hss$\bullet$\hskip 0.5em}A charity allocation task, where you will help to disburse funds to real-world charities.

\smallskip
\textbf{Compensation \& Impact.} You will be paid a base rate of \$15 USD, with
potential to earn a bonus depending on your performance and decisions in the
three tasks.

\textbf{Confidentiality.} The results of this study will be used solely for
research purposes. Our team will keep all your information from this study
strictly confidential, as required by law.

\textbf{Voluntary Participation.} Your participation is voluntary, which means
you can choose whether or not to participate. You may choose not to participate
by exiting the task at any point; \textbf{however, your submission will only be
accepted if you complete the entire task in its entirety.} There are no known
costs to you for participating in this research study except for your time.

By checking the box below and proceeding, you are acknowledging that you are
over the age of 18 and that you consent to participate.
\end{quote}

\paragraph{Payment structure.}
Every participant received a \textbf{\$15 USD base payment} for the session,
which lasted approximately one hour (three $5$--$10$ minute discussions plus
the intake, comprehension, survey, and exit stages of \S\ref{ssec:app_flow}).
On top of the base payment, three task-contingent components were announced
to participants before the corresponding task:
\begin{compactitem}

  \item \textbf{Open-ended discussion bonus (\$2).} A participant earned a
        \$2 bonus if they contributed at least six substantive messages
        during the discussion, each longer than five words. This is the
        criterion scored as ``Engagement qualified'' in Table~\ref{tab:metrics};
        $22.4\%$ of Study~1 and $20.1\%$ of Study~2 participant-episodes
        qualified (Appendix~\ref{ssec:app_outcomes}).
  \item \textbf{Negotiation bonus (up to \$10 per triad).} Each member of the
        agreed coalition received, as a bonus, the dollar amount allocated to
        them in the agreed split, provided that every coalition member's
        privately recorded split matched (\S\ref{ssec:app_tasks}). A
        participant left out of the coalition, or any member of a coalition
        whose private records disagreed, received \$0 from this task.
  \item \textbf{Charity donation (\$3, not paid to participants).} For the
        consensus task, the experimenters donated \$3 to the charity the group
        selected, or to the charity drawn at random when the group agreed to
        disagree. This is an other-regarding stake: it gives the decision
        real consequences without changing the participant's own payment.
\end{compactitem}
Participants were told that payment would be reviewed and released through
Prolific within 24--48 hours; the open-ended task introduction stated 48--72
hours for its bonus.

\paragraph{Debriefing.}
A debriefing stage disclosed the research purpose verbatim as ``to understand how
AI facilitator helps different human conversation tasks,'' repeated the
compensation terms, and pointed participants to the research team via Prolific.
Participants were informed on the study-overview stage, before any task, that
``an AI facilitator will join your session''; the facilitator was never presented
as human, and the facilitator's own instructions in both studies forbade claiming
otherwise.

\subsection{Participant Flow}
\label{ssec:app_flow}

The experiment comprised 38 stages. Each participant passed through an intake
block, then the three task blocks in a counterbalanced order, then an exit block.

\begin{compactitem}
\setlength{\itemsep}{1pt}
  \setlength{\parskip}{0pt}
  \item \textbf{Intake.} Platform Tutorial; Terms of Service (consent gate);
        View your Profile (anonymous animal-colour alias and avatar); Study
        Overview; \emph{Survey on AI Facilitation} (baseline attitudes, measured
        \emph{before} any exposure); Transfer (wait up to $10$ minutes until a
        cohort of three is assembled).
  \item \textbf{Task block}, repeated three times in the cohort's assigned
        order: task introduction (two info stages for negotiation);
        comprehension check; a role-assignment stage for negotiation only;
        pre-discussion survey; the \textbf{chat stage} ($5$--$10$ minutes,
        facilitator present); then, for the consensus and negotiation tasks, a
        private final-decision stage followed by an outcome reveal; and finally
        the post-discussion survey, the group questionnaire and the AI
        facilitator evaluation.
  \item \textbf{Exit.} \emph{Survey: About You} (profession and three free-text
        self-disclosure items); \emph{Survey: About Others} (one free-text
        impression per co-participant); Debriefing; \emph{Survey on Experiment
        Feedback}; Experiment End.
\end{compactitem}

Note that the outcome reveal precedes the post-discussion survey, so
satisfaction and psychological-safety ratings are given in knowledge of the
task result, and that the debriefing precedes the final feedback survey.

\paragraph{Facilitator implementation.}
The facilitator referenced in \S\ref{sec:llm_facilitation} runs as an
asynchronous, non-blocking loop on the platform. \emph{Concurrent generation:}
on every incoming participant message the facilitator is consulted with the
updated transcript and drafts a candidate decision while participants keep
typing. \emph{Context invalidation:} if further participant messages arrive
before the draft is sent, the pending draft is discarded and the facilitator is
consulted again against the freshest transcript, so no message is ever posted
on a stale view of the conversation. \emph{Unbiased deliberation:} the
facilitator sees only the task introduction, the stage description, the
participant roster (anonymous alias and avatar, or party letter in
negotiation), and the live transcript; it never receives persona, demographic,
or survey information. Every consultation returns the JSON decision of
Appendix~\ref{ssec:app_prompt_facilitator}---a timing rationale, a speak/silent
flag, the message, and the cited playbook rules---and the platform logs it
whether or not the facilitator spoke. The scaffold is identical in both
studies; only the block that carries the expert prompt (Study~1) or the learned
playbook (Study~2) differs.

\paragraph{Counterbalancing.}
Task order was counterbalanced across the six permutations of $\tasks$
(\texttt{OCN}, \texttt{ONC}, \texttt{CNO}, \texttt{CON}, \texttt{NCO},
\texttt{NOC}; O~=~open-ended, C~=~consensus, N~=~negotiation), so that task
identity is orthogonal to serial position. Assignment is at the
\emph{condition} level---each permutation was fielded as a separate recruitment
with its own stage ordering---rather than randomised per cohort within one
recruitment (\S\ref{ssec:app_equivalence}). Both studies realised all six orders
in near-equal proportion (Table~\ref{tab:app_sample}).

\subsection{Task Materials}
\label{ssec:app_tasks}

All three tasks run in a synchronous text channel with three participants and the
facilitator, under a $10$-minute cap with a $5$-minute minimum
(\texttt{timeLimitInMinutes}~$=10$, \texttt{timeMinimumInMinutes}~$=5$); after
five minutes participants may advance early. Task materials are byte-identical
between Study~1 and Study~2 (\S\ref{ssec:app_equivalence}).

\subsubsection{Task 1: Open-Ended Discussion}

The group discusses a policy motion with no requirement to agree. The motion is
drawn per cohort from a pool of \textbf{two}, by a seeded random permutation:

\begin{quote}\small
\textbf{(fr)} Local law enforcement agencies should be allowed to use live
facial recognition technology in public spaces.

\textbf{(medicaid)} The federal government should mandate that anyone who
previously qualified for Medicaid under the Affordable Care Act of 2010 needs to
work, do community service or go to school to retain their eligibility.
\end{quote}

Realised allocation: \texttt{fr} in $74$ (S1) and $79$ (S2) cohorts,
\texttt{medicaid} in $70$ and $75$. The task introduction announced the \$2
engagement bonus and its criterion (\S\ref{ssec:app_recruit}). Because the pool
contains only two motions---both US-centric and politically charged---the
open-ended results should be read as evidence about facilitation under
contested-values discussion, not as broad topical generalisation.

\begin{table}
\centering

\small
\setlength{\tabcolsep}{4pt}
\begin{tabularx}{\linewidth}{@{}lLll@{}}
  \toprule
  \textbf{Measure} & \textbf{What it asks, or how it is computed} & \textbf{Scale} & \textbf{Role} \\
  \midrule
  \multicolumn{4}{@{}p{\linewidth}@{}}{\emph{(1) Group experience $\mathbf{s}_e$: how the participants felt, self-reported after every task}} \\
  \textbf{Psychological safety} & ``I felt psychologically safe during this discussion.'' & 1--5 $\uparrow$ & \textbf{Primary} \\
  \quad \textit{spread} $\sigma_{\text{w}}$ & Disagreement among the three members of one episode on the item above; low means the conversation felt the same to everyone. & $\geq 0$ $\downarrow$ & \textbf{Primary} \\
  \addlinespace
  Felt heard & ``I felt heard and understood during the discussion.'' & 1--5 $\uparrow$ & Secondary \\
  Comfortable speaking & ``I felt comfortable speaking up and voicing my opinions.'' & 1--5 $\uparrow$ & Secondary \\
  Friction and anger & ``There was friction and anger between the participants.'' & 1--5 $\downarrow$ & Secondary \\
  Group productive & ``The discussion was engaging and productive.'' & 1--5 $\uparrow$ & Secondary \\
  Sensitive disclosure & ``The participants revealed sensitive personal information or feelings.'' Neither direction is preferable; reported for context only. & 1--5 --- & Descriptive \\
  \midrule
  \multicolumn{4}{@{}p{\linewidth}@{}}{\emph{(2) Facilitator evaluation $\mathbf{f}_e$: how the participants rated the AI facilitator}} \\
  Facilitator productive & ``The AI facilitator made the group discussion more productive.'' & 1--5 $\uparrow$ & Secondary \\
  Facilitator satisfying & ``I was satisfied with the performance of the AI facilitator.'' & 1--5 $\uparrow$ & Secondary \\
  Comfort with facilitator & ``I felt comfortable having the AI facilitator in the group discussion.'' & 1--5 $\uparrow$ & Secondary \\
  Prior-calibrated shift & Each item above minus the \emph{same} participant's pre-study expectation of AI facilitation, collected before any discussion ($\text{Post}-\text{Pre}$); separates the facilitator's effect from the participant's prior attitude toward AI. & $-4$ to $+4$ $\uparrow$ & Secondary \\
  \midrule
  \multicolumn{4}{@{}p{\linewidth}@{}}{\emph{(3) Task outcome $\omega_e$: what the group achieved; one value per episode unless marked per participant}} \\
  Unanimous consensus & $\tau_{\text{cons}}$: all three participants settle on the same charity. & $\{0,1\}$ $\uparrow$ & Secondary \\
  Coalition agreed & $\tau_{\text{neg}}$: the group closes a deal naming a winning coalition. & $\{0,1\}$ $\uparrow$ & Secondary \\
  Valid allocation & $\tau_{\text{neg}}$: the agreed shares sum exactly to the available pot. & $\{0,1\}$ $\uparrow$ & Secondary \\
  Trust in partners & $\tau_{\text{neg}}$, per participant: reported trust in the co-negotiators after the deal. & 1--5 $\uparrow$ & Secondary \\
 
  Message volume & $\tau_{\text{open}}$, per participant: substantive messages sent during the discussion (Table~\ref{tab:live}). & count $\uparrow$ & Secondary \\
  Engagement qualified & $\tau_{\text{open}}$, per participant: sent at least six substantive messages, the bonus criterion announced to participants in advance. & $\{0,1\}$ $\uparrow$ & Secondary \\
  
  \bottomrule
\end{tabularx}
\caption{\textbf{Every measure we report, and which one is primary.} $\uparrow$/$\downarrow$ mark the preferred direction; ``---'' means no direction is preferred. Psychological safety is the single primary metric and is read both as a level and as a within-group spread; all other survey measures are secondary and Benjamini--Hochberg corrected within their block. Facilitator turns and words (Table~\ref{tab:live}) are diagnostic: fewer is better only \emph{at equal experience and outcome}, since they measure the cost of facilitation, not its quality. Verbatim item wording and response anchors are in Table~\ref{tab:app_items}.}
\label{tab:metrics}
\end{table}
\subsubsection{Task 2: Consensus-Building (Charity Allocation)}

The group must reach \emph{unanimous} agreement on which single charity receives
a \$3 donation made by the experimenters; a stable disagreement resolves by
uniform random draw. We do not penalize disagreement by forfeiting the donation since we do not want humans to feel this can lead to a very negative impact. We hope humans can make decision without much pressure.  Three charities are shown, drawn as the first three of a
per-cohort seeded random permutation of the nine in Table~\ref{tab:app_charities}.
Slot position is also randomised. Across the $298$ cohorts this realised $251$
distinct ordered triples and $83$ of the $84$ possible unordered $3$-sets, with
per-charity exposure close to the uniform expectation of $99.3$ appearances
(range $83$--$116$). Participants saw each charity's name, mission and
Charity~Navigator link; the \texttt{score} field present in the configuration was
\emph{not} rendered in any stage, so participants did not see charity ratings.

\begin{table}[t]
\centering
\scriptsize
\setlength{\tabcolsep}{4pt}
\begin{tabularx}{\linewidth}{@{}lLr@{}}
\toprule
\textbf{Charity} & \textbf{Mission (verbatim)} & \textbf{Shown} \\
\midrule
Sudan Humanitarian Aid & To provide life-saving aid to the affected population, Sadagaat-USA is collaborating with other US-based organizations and local initiatives in Sudan to offer food, medication, medical supplies, and water through its emergency response program. & $116$ \\
International Fund for Animal Welfare & Fresh thinking and bold action for animals, people, and the place we call home. & $102$ \\
WildAid & WildAid's mission is to end the illegal wildlife trade in our lifetimes by reducing demand through public awareness campaigns and providing comprehensive marine protection. & $101$ \\
Aid for Children in Remote Villages & [Facilitated via GlobalGiving] The Eden Social Welfare Foundation has cared for underprivileged children since 2006, with the hope that they can enjoy the right to a fair education, better after-school care, and a healthy and nutritious breakfast. & $99$ \\
Global Fund for Women & Global Fund for Women advances women's human rights by investing in women-led organizations worldwide. Our international network of supporters mobilizes financial and other resources to support women's actions for social justice, equality and peace. & $99$ \\
Clean Ocean Action & Clean Oceans International is dedicated to reducing plastic pollution in the world's ocean through Research, Innovation, and Direct Action. & $99$ \\
Eyecare in India & Our mission is to reach out to the rural poor and provide quality eye care free of cost to the needy by building operationally self-sufficient super specialty eye care hospitals across India and perform free eye surgeries. & $99$ \\
Global Housing for Orphans & Givelight builds nurturing homes and provides high quality education for orphans globally. & $96$ \\
Rainforest Action & Rainforest Action Network campaigns for the forests, their inhabitants and the natural systems that sustain life by transforming the global marketplace through education, grassroots organizing and non-violent direct action. & $83$ \\
\bottomrule
\end{tabularx}
\caption{The nine-charity pool. ``Shown'' counts appearances among the three
presented, pooled over the $298$ cohorts of both studies ($894$ slot-fills,
uniform expectation $99.3$).}
\label{tab:app_charities}
\end{table}

\subsubsection{Task 3: Multi-Party Negotiation}

Participants are randomly assigned roles A, B or C and negotiate which coalition
forms and how to divide its pot:

\begin{center}\small
\begin{tabular}{@{}lrl@{}}
\toprule
\textbf{Coalition} & \textbf{Pot} & \textbf{Excluded party} \\
\midrule
A + B     & \$9  & C receives \$0 \\
A + C     & \$7  & B receives \$0 \\
B + C     & \$5  & A receives \$0 \\
A + B + C & \$10 & --- \\
Going solo & \$0 & --- \\
\bottomrule
\end{tabular}
\end{center}

Shares must be integers, enforced by a $0$--$10$ slider with unit steps. After
the discussion each participant \emph{privately} records the agreed split; if
coalition members' records disagree, nobody is paid. This makes explicit
arithmetic agreement, not merely verbal assent, the operative success condition,
and it is the mechanism behind the deal-validity outcome in
Table~\ref{tab:outcomes}.

\paragraph{No informational asymmetry.}
The asymmetry between roles is purely \emph{structural}---A belongs to the two
highest-value pairs---and not informational. The full payoff table is shown to
every participant in the task introduction, the comprehension check, the
pre-negotiation survey, the chat stage and the final decision, and the three role
profiles are identical apart from the party letter. Role assignment was balanced
in both studies (S1 $145/145/143$; S2 $154/154/155$).



\subsubsection{Protocol Equivalence Between Studies}
\label{ssec:app_equivalence}

Because Study~2 is a between-study comparison (\S\ref{ssec:study2}), the claim
that the facilitator prompt is the \emph{sole} manipulation needs verifying
rather than asserting. We diffed all $12$ study configurations---six
counterbalancing conditions $\times$ two studies---keying stages by their stable
identifier and comparing full JSON with the positional \texttt{name} field
removed.

All $38$ stage identifiers are present in all $12$ configurations, and
\textbf{for every one of the six orders the Study~1 and Study~2 configurations
are identical}: same stage sequence, same task introductions, same motion and
charity pools with the same seeded selection mechanism, same payoff structure and
role profiles, same comprehension items, and same survey instruments. Only three
stages vary anywhere in the corpus, and each varies \emph{across counterbalancing
conditions rather than across studies}: the negotiation comprehension check
(four variants, including the defect above), the negotiation introduction (an
escaped-markdown rendering difference in \texttt{ONC}), and a
\texttt{waitForAllParticipants} flag on the payout summary (also \texttt{ONC}).

\paragraph{Counterbalancing is between-condition, not within-study.}
One design detail deserves precision. The six task orders were not randomised
within a single recruitment; each order was fielded as its own Prolific study
with its own stage ordering. Order is therefore orthogonal to task identity by
construction, but it is confounded with recruitment batch. Both studies realised
all six orders in near-equal proportion (Table~\ref{tab:app_sample}), so this is
balanced across the Study~1 versus Study~2 contrast.

\subsection{Survey Instruments}
\label{ssec:app_surveys}

Table~\ref{tab:app_items} lists every closed-form item, its response scale and
its anchors, verbatim. All Likert items are five-point unless noted; the two
experiment-feedback items are seven-point; two negotiation items are 0--10
sliders. Free-text items are listed in the caption rather than the body.

\begin{table}[t]
\centering
\footnotesize
\setlength{\tabcolsep}{4pt}
\begin{tabularx}{\linewidth}{@{}llL>{\raggedright\arraybackslash}p{2.7cm}@{}}
\toprule
\textbf{Stage} & \textbf{Variable} & \textbf{Item (verbatim)} & \textbf{Scale / anchors} \\
\midrule
\multicolumn{4}{@{}l}{\emph{Intake, once per participant, before any task (\texttt{ai\_facilitation\_pre\_survey})}}\\
AI Facilitation & \texttt{pre\_ai\_used} & I have used AI assistants for interpersonal tasks, such as writing messages or resolving conflicts. & 1--5 Rarely / Frequently \\
AI Facilitation & \texttt{pre\_fac\_productive} & I believe an AI facilitator could make group discussions more productive. & 1--5 SD / SA \\
AI Facilitation & \texttt{pre\_fac\_comfort} & I would feel comfortable having an AI facilitator in the group discussion. & 1--5 SD / SA \\
AI Facilitation & \texttt{pre\_fac\_willing} & If given the option, I would be willing to use an AI facilitator in group discussions. & 1--5 SD / SA \\
\midrule
\multicolumn{4}{@{}l}{\emph{Pre-discussion, per task}}\\
Task 1 pre & --- & What is your initial stance on this motion? & 1--5 SD / SA \\
Task 2 pre & --- & Take this time to learn more about the three charities. After the reading, what's your initial preference? & 3-option MC \\
Task 2 pre & --- & How strongly do you feel about your decision? & 1--5 Not strong at all / Very strong \\
Task 3 pre & --- & In your ideal coalition, exactly how much money do you want to secure for yourself? & 0--10 slider \\
Task 3 pre & --- & What is your ``Walk-Away'' point? & 0--10 slider \\
\midrule
\multicolumn{4}{@{}l}{\emph{Post-discussion, per task}}\\
Post & --- & After discussing with your group, what is your final stance on this motion? \textit{(Task 1)} & 1--5 SD / SA \\
Post & --- & I feel strongly about my final allocation. \textit{(Task 2)} & 1--5 SD / SA \\
Post & --- & I am satisfied with my final outcome. \textit{(Task 3)} & 1--5 SD / SA \\
Post & \texttt{felt\_heard} & I felt heard and understood during the discussion. & 1--5 SD / SA \\
Post & \texttt{comfort\_speaking} & I felt comfortable speaking up and voicing my opinions. & 1--5 SD / SA \\
Post & \texttt{psych\_safety} & I felt psychologically safe during this discussion. & 1--5 SD / SA \\
\midrule
\multicolumn{4}{@{}l}{\emph{Group questionnaire, per task}}\\
Group & \texttt{friction} & There was friction and anger between the participants. & 1--5 SD / SA \\
Group & \texttt{sensitive}$^\dagger$ & The participants revealed sensitive personal information or feelings. & 1--5 SD / SA \\
Group & \texttt{group\_productive} & The discussion was engaging and productive. & 1--5 SD / SA \\
\midrule
\multicolumn{4}{@{}l}{\emph{AI facilitator evaluation, per task}}\\
Facilitator & \texttt{fac\_productive} & I believe that the AI facilitator made the group discussion more productive. & 1--5 SD / SA \\
Facilitator & \texttt{fac\_comfort} & I felt comfortable having the AI facilitator in the group discussion. & 1--5 SD / SA \\
Facilitator & \texttt{fac\_satisfied} & I was satisfied with the performance of the AI facilitator. & 1--5 Strongly Dissatisfied / Strongly Satisfied \\
\midrule
\multicolumn{4}{@{}l}{\emph{Exit, once per participant}}\\
About You & \texttt{profession} & What best describes your current profession? & 10-option MC \\
Feedback & \texttt{exp\_overall} & Overall, how would you rate your experience in this study? & 1--7 Very Negative / Very Positive \\
Feedback & \texttt{exp\_clarity} & How clear were the instructions and questions throughout the experiment? & 1--7 Very Unclear / Very Clear \\
\bottomrule
\end{tabularx}
\caption{Complete closed-form survey instrument. SD means ``strongly disagree'' while SA means ``strongly agree''.}
\label{tab:app_items}
\end{table}

\subsection{Realised Sample}
\label{ssec:app_sample}

\begin{table}[t]
\centering
\footnotesize
\setlength{\tabcolsep}{5pt}
\begin{tabular}{@{}lrr@{}}
\toprule
& \textbf{Study 1 (expert)} & \textbf{Study 2 (learned)} \\
\midrule
Participants                           & $432$ & $462$ \\
Cohorts                                & $144$ & $154$ \\
Episodes                               & $432$ & $462$ \\
Participant-episode records            & $1302$ & $1389$ \\
\midrule
\multicolumn{3}{@{}l}{\emph{Task order (cohorts)}}\\
\texttt{OCN} / \texttt{ONC}            & $24$ / $24$ & $27$ / $26$ \\
\texttt{CNO} / \texttt{CON}            & $23$ / $26$ & $23$ / $26$ \\
\texttt{NCO} / \texttt{NOC}            & $22$ / $25$ & $26$ / $26$ \\
\bottomrule
\end{tabular}
\caption{Realised analytic sample: cohorts that completed all three tasks.}
\label{tab:app_sample}
\end{table}

\paragraph{Analytic sample.}
All analyses use cohorts that completed all three tasks: $144$ cohorts
($432$ participants, $432$ episodes) in Study~1 and $154$ cohorts ($462$
participants, $462$ episodes) in Study~2, the figures used throughout
\S\ref{sec:results}.

\subsection{Participant Background}
\label{ssec:app_demog}

\begin{table}[t]
\centering
\footnotesize
\setlength{\tabcolsep}{5pt}
\begin{tabular}{@{}lrr@{}}
\toprule
\textbf{Occupational category} & \textbf{Study 1} & \textbf{Study 2} \\
\midrule
Computer, Engineering, \& Science                 & $19.6\%$ & $19.7\%$ \\
Non-Employed (Student, Retired, Homemaker, Seeking Work) & $16.2\%$ & $17.5\%$ \\
Management, Business, \& Financial Operations     & $15.2\%$ & $18.8\%$ \\
Service, Sales, \& Hospitality                    & $15.0\%$ & $10.8\%$ \\
Healthcare Practitioners \& Support               & $10.6\%$ & $9.5\%$ \\
Arts, Design, Media, \& Entertainment             & $7.2\%$  & $6.0\%$ \\
Education, Training, \& Library                   & $5.5\%$  & $6.3\%$ \\
Trades, Construction, \& Manufacturing            & $4.4\%$  & $6.5\%$ \\
Government, Law, \& Public Safety                 & $3.2\%$  & $3.0\%$ \\
Farming, Maintenance, \& Specialized Outdoor      & $2.1\%$  & $0.9\%$ \\
Not reported                                      & $0.9\%$  & $1.1\%$ \\
\bottomrule
\end{tabular}
\caption{Self-reported occupation, the only structured demographic collected.
Percentages are of unique participants within each study.}
\label{tab:app_demog}
\end{table}

\paragraph{What was and was not collected.}
We did not collect participant-identifying information across both empirical studies; the \emph{only} structured demographic data collected is the
ten-category occupational item in Table~\ref{tab:app_demog}. \textbf{Age, gender,
race or ethnicity, education, income, and country were not collected}; the released records carry a Prolific
identifier, an anonymous alias, and an avatar. Participant background is
otherwise represented only by three free-text self-disclosure prompts on the
\emph{About You} stage: a description of self-described personality attributes; one to three regular hobbies; and a reflection on which personal values
or experiences influenced the participant's choices in the study. This limits
any subgroup analysis, and we make no claims about representativeness beyond the
Prolific pool.

Consistent with the facilitator design (\S\ref{sec:llm_facilitation}), none of
this information was ever visible to the facilitator, which saw only the task
instructions and the live transcript.

\paragraph{Baseline attitudes toward AI facilitation.}
Table~\ref{tab:app_attitudes} reports the four intake items, measured before any
exposure. The two samples are close but not identical: Study~2 participants
arrived slightly \emph{less} enthusiastic on all three evaluative items. The
largest gap, $0.21$ points on ``AI facilitator could make discussions more
productive,'' is about two-thirds the size of the $-0.30$ raw post-task deficit in
perceived facilitator productivity (Table~\ref{tab:app_raw_fac}). Because
Study~2 is a between-study comparison (\S\ref{ssec:study2}), this prior
difference is a genuine alternative explanation for part of that deficit;
\S\ref{ssec:live_results} and Appendix~\ref{ssec:app_polarization} quantify it
by calibrating each post-task rating against the same participant's pre-study
response.

\begin{table}[t]
\centering
\footnotesize
\setlength{\tabcolsep}{5pt}
\begin{tabular}{@{}lrrr@{}}
\toprule
\textbf{Intake item (1--5)} & \textbf{Study 1} & \textbf{Study 2} & $\Delta$ \\
\midrule
Used AI assistants for interpersonal tasks        & $3.38 \pm 1.37$ & $3.32 \pm 1.39$ & $-0.06$ \\
AI facilitator could make discussions more productive & $4.00 \pm 0.92$ & $3.79 \pm 1.05$ & $-0.21$ \\
Comfortable having an AI facilitator              & $4.18 \pm 0.89$ & $4.05 \pm 0.99$ & $-0.13$ \\
Willing to use an AI facilitator                  & $4.11 \pm 0.93$ & $3.95 \pm 1.04$ & $-0.16$ \\
\midrule
\textbf{Exit item (1--7)} & & & \\
Overall experience in this study                  & $6.47 \pm 0.96$ & $6.45 \pm 1.05$ & $-0.02$ \\
Clarity of instructions and questions             & $6.68 \pm 0.66$ & $6.66 \pm 0.78$ & $-0.02$ \\
\bottomrule
\end{tabular}
\caption{Pre-exposure attitudes and exit evaluations, de-duplicated to one
response per participant. Exit ratings are near-ceiling and indistinguishable across
studies, indicating the Study~2 deficit is specific to the facilitator rather
than a general degradation of the session. Means here are per participant and
therefore differ in the second decimal from the participant-task-level figures
of Appendix~\ref{ssec:app_polarization}, where each pre-study response is paired
with that participant's three post-task ratings.}
\label{tab:app_attitudes}
\end{table}

\subsection{Denominators}
\label{ssec:app_discrepancies}

Every count in \S\ref{sec:results} and in this appendix is stated on the
analysis basis: $432$ episodes from $144$ cohorts in Study~1 and $462$ episodes
from $154$ cohorts in Study~2, i.e.\ cohorts completing all three tasks, the
same criterion in both studies. The survey, behavioral, and outcome rows
therefore share a common denominator. Retention was, if anything, higher in
Study~2 than in Study~1, so differential attrition does not explain the raw
facilitator-rating deficit (Appendix~\ref{ssec:app_robustness}).


\section{Learning Algorithm Details}
\label{sec:appendix_method}

This appendix gives the full specification of the components summarized in
\S\ref{sec:method}: how the Study~1 corpus is recorded and split
(\S\ref{ssec:app_corpus}), how the playbook parameterizes the policy and how
decision points are sampled (\S\ref{ssec:app_playbook}), how the judge is
constructed and aligned to human annotators (\S\ref{sec:appendix_rubrics}), and
the reflector, curator, credit-attribution, and pruning machinery of the
optimization loop (\S\ref{ssec:app_loop}).

\subsection{Corpus Construction and Splits}
\label{ssec:app_corpus}

\paragraph{Procedure.}
Each episode $e = (g_j, \tau, k)$ places cohort $g_j$ in task $\tau$ at round
$k \in \{1,2,3\}$, moderated by an autonomous LLM facilitator operating under the
asynchronous, non-blocking protocol of \S\ref{sec:llm_facilitation}. A discussion
stage unlocks after $5$ minutes and closes automatically at $10$ minutes.
Participants exchange $26.9$ messages per episode on average ($21.6$ participant
messages and $5.3$ facilitator messages including the scripted welcome;
Table~\ref{tab:live}), and the
span from the first to the last chat message averages $6.6$ minutes (median
$6.1$). Immediately after each episode, every participant completes the
instruments of \S\ref{sec:eval_metrics}: Likert items for psychological safety
$\mathbf{s}_e$, facilitator satisfaction $\mathbf{f}_e$, and task-specific
outcome ratings, each accompanied by free-text rationales. Objective outcomes
$\omega_e$ are logged by the platform. Recruitment, consent, compensation, task
materials, and the full survey instruments are documented separately in
Appendix~\ref{sec:appendix_design}.

\paragraph{Episode records.}
Each episode yields the tuple
\begin{equation}
    \eprecord_e = \bigl(x_e,\; c_e,\; q_e,\; o_e\bigr),
\end{equation}
where $x_e = (m_1,\ldots,m_L)$ is the chronologically ordered message sequence
(each $m_i$ carrying a speaker role $\in \{\textsc{Facilitator},
\textsc{Participant}, \textsc{System}\}$, sender identity, second-level
timestamp, and text); $c_e$ is the task context shown to participants; $q_e$
collects the post-task survey responses and free-text rationales of all three
participants; and $o_e = (\mathbf{f}_e, \mathbf{s}_e, \omega_e)$ is the
evaluation tuple of \S\ref{sec:eval_metrics}. The platform additionally logs, at
every turn where the facilitator was consulted, whether it spoke and what it
said. We use this logged \emph{action} as the reference move. Its recorded reasoning
never enters the learning loop: the reflectors and the curator never see it,
and the judge inside the loop compares two learned-policy outputs rather than
the expert move (\S\ref{ssec:policy}), so the expert prompt's idiosyncratic
self-justifications cannot be distilled into the playbook. Only in the offline
benchmark, after learning has finished, is the logged expert move shown to the
judge together with its recorded explanation (\S\ref{sec:appendix_rubrics}).
The resulting corpus is $\dataset = \{\eprecord_e\}$ with
$|\dataset| = 432$.

\paragraph{Splits.}
We partition $\dataset$ at the \emph{cohort} level, stratified by task, in a
$7{:}1{:}2$ ratio. All three episodes of a cohort land in the same split, so no
group's dynamics leak across splits: $\dataset_{\text{train}}$ ($102$ cohorts,
$306$ episodes), $\dataset_{\text{val}}$ ($13$ cohorts, $39$ episodes), and
$\dataset_{\text{test}}$ ($29$ cohorts, $87$ episodes), the last untouched until
final evaluation.

\subsection{Playbook Representation and Decision Points}
\label{ssec:app_playbook}

\paragraph{Rule structure.}
Each rule $c \in \playbook$ carries a stable identifier, a section, its
instruction text, and four counters: $\text{helpful}(c)$, $\text{harmful}(c)$,
$\text{uses}(c)$ (lifetime citations, regardless of verdict), and
$\text{unused}(c)$ (consecutive generations without a citation). Sections
follow a fixed eight-part \emph{cue schema}, and each identifier is the
section's slug followed by a five-digit serial: \texttt{obj} (Role and
Objective), \texttt{cri} (Success Criteria), \texttt{tmg} (Time Management \&
Pacing), \texttt{gui} (General Guidelines), \texttt{fm} (Recognize Failure
Modes), \texttt{pri} (Core Principles), \texttt{res} (Core Responsibility),
and \texttt{donot} (Do NOT do the following). Serials are never reused, so a
rule keeps its identifier through edits and merges, and every identifier cited
in this appendix resolves to exactly one rule of the deployed playbook
reproduced in Appendix~\ref{ssec:app_playbook_text}. Because rules are discrete
and individually addressable, credit and blame can be assigned at rule
granularity, and the resulting artifact is human-readable and auditable.

\paragraph{Sampling the evaluation set.}
Deciding when to remain silent is as consequential as deciding what to say, so
we evaluate on a balanced set. From each episode we take every step at which the
deployed expert facilitator spoke, excluding the scripted welcome
($\mathcal{I}_e^+$), and an equal number of steps at which it stayed silent,
sampled by uniform stride across the episode ($\mathcal{I}_e^-$), giving
$\mathcal{I}_e = \mathcal{I}_e^+ \cup \mathcal{I}_e^-$ with a $1{:}1$
speech/silence ratio and logged reference actions $a_i^{\text{ref}}$. This
yields $8.9$ decision points per episode on average ($4.4$ spoken, $4.4$ silent)
and $3{,}730$ across the corpus.

The $1{:}1$ balance is a property of the \emph{benchmark}, not of natural
conversation; together with the mutual-silence fast path
(\S\ref{sec:appendix_rubrics}) it makes abstention a low-risk action, which is
the limitation discussed at the end of the main text.

\subsection{Judge Design and Human Alignment}
\label{sec:appendix_rubrics}

\paragraph{Pairwise judging.}
The judge scores each action on $\textsf{timing} \in \{1..5\}$ and, for speech,
$\textsf{content} \in \{1..5\}$, with $\textsf{overall} = (\textsf{timing} +
\textsf{content})/2$ for \textsc{Speak} and $\textsf{overall} = \textsf{timing}$
for \textsc{Silent}. A post-hoc consistency pass enforces that the stated
verdict agrees with the numeric scores---equal overall scores are coerced to a
tie---so the verdict cannot drift from the evidence the judge itself recorded.
Presentation order is randomized by a deterministic hash of the context and both
actions, eliminating position bias while keeping runs reproducible, and all
judge calls use temperature $0$. Each reaction is shown with the reasoning it
carries, and the rubric instructs the judge that a rationale is a claim to be
checked against the transcript rather than evidence, that a well-argued
rationale earns no credit on its own, and that the absence of one is not a
flaw. During learning, both reactions are learned-policy outputs
($\pi_{\widetilde{\playbook}_t}$ versus $\pi_{\playbook_{t-1}}$,
\S\ref{ssec:policy}), each carrying its own rationale $\rho_i$. Only at test
time (\S\ref{ssec:offline}) is the logged expert move $a_i^{\text{ref}}$ the
competitor; it is then shown with the explanation the Study~1 facilitator
recorded alongside it. When both facilitators stay silent, the harness records
a tie without calling the judge (the \emph{mutual-silence fast path}). The
verbatim rubric is reproduced in Appendix~\ref{ssec:app_prompt_judge}.

\paragraph{Why alignment is necessary.}
An off-the-shelf judge is not an acceptable signal. Our first rubric agreed with
human annotators at only Cohen's $\kappa = 0.29$ and, when used for learning,
wrote its own pathologies into the playbook---most conspicuously a rule
mandating an intervention every $K$ participant turns, a cadence no human
annotator endorsed. Optimizing against an unaligned judge does not merely add
noise; it installs the judge's biases as explicit policy.

\paragraph{Human annotation protocol.}
We sampled $50$ decision points from the benchmark ($42$ speech, $8$ silence)
and recruited $20$ human annotators. Each annotator judged $10$ points and each
point received $4$ independent votes, for $200$ votes in total. Annotators
worked in a purpose-built interface that showed exactly what the judge
sees---task instructions, the transcript prefix with elapsed timestamps, and the
two fully blinded candidate reactions with their rationales---and recorded a
preference plus free-text notes. 

To establish the ground-truth human reference, we aggregate the four votes per
decision point via majority rule: the option receiving at least three votes is
designated as the preferred reaction. In cases where the panel is split evenly
(two votes for candidate A and two votes for candidate B), the decision point is
formally treated as a tie.

\paragraph{LLM judge Rubric refinement.}
Reflecting over agreement and disagreement cases revealed that annotators were
implicitly conditioning on the \emph{state of the exchange} before evaluating
any action. We restructured the rubric to make this explicit and two-stage:
\begin{compactitem}
\setlength{\itemsep}{1pt}
  \setlength{\parskip}{0pt}
    \item \textbf{State diagnosis.} Before inspecting either candidate, the
    judge must classify the conversation into exactly one of six observable
    states---\textsc{Productive}, \textsc{Opening/Shallow}, \textsc{Confused},
    \textsc{Stalled}, \textsc{Unproductive}, \textsc{Settled}---and quote the
    evidence for it. Timing is then scored against \emph{what that state calls
    for}, not against how long the facilitator has been quiet.
    \item \textbf{Binding content caps.} Content is scored from a ceiling of $5$
    downward through non-compensable caps: a claim that misdescribes the
    discussion caps content at $2$; endorsing a stance or declaring an unmade
    decision caps at $3$; an unclear addressee caps at $4$; brusqueness toward a
    participant caps at $3$.
\end{compactitem}

\paragraph{Alignment attained.}
The aligned judge reaches Cohen's $\kappa = 0.70$ against the panel majority
($80\%$ raw agreement over the $50$ points, and $100\%$ on points with unanimous
or $3$--$1$ human majorities), and matches an individual human vote $66\%$ of the
time---above the $47\%$ human--human baseline. We stress the interpretation: the
judge is a reliable proxy for the \emph{median} annotator, not an oracle.

\subsection{Reflectors, Curator, and Credit Attribution}
\label{ssec:app_loop}

This section specifies the components of Figure~\ref{fig:algo}. The verbatim
prompt of every component is reproduced in Appendix~\ref{sec:appendix_prompts};
decision-point sampling is given in \S\ref{ssec:app_playbook} and the judge in
\S\ref{sec:appendix_rubrics}.

\paragraph{Conversation reflector (human evidence).}
At generation $g$, this agent reads a batch $\mathcal{B}_g \subset
\dataset_{\text{train}}$ of full episode records: transcripts $x_e$, participant
surveys $q_e$, and outcomes $o_e$. It cross-references each approving or
critical participant remark with the facilitator turns that provoked it, and
emits transferable lessons $L^{\text{conv}}$ keyed to \emph{observable
triggers}. It answers ``what do participants reward?'' and never inspects a
model rollout.

\paragraph{Validation reflector (self-critique).}
This agent reads only the judged validation turns of the current policy.
Analyzing losses ($v_i = \text{win}_{g-1}$) together with the cited rules $C_i$
and the judge's rationale $r_i$, it classifies breakdowns as \emph{rule misfire}
(a trigger fired in an inapplicable context), \emph{overreach} (valid trigger,
excessive intervention), \emph{rule conflict}, or \emph{uncovered state}. To
prevent overfitting to idiosyncratic turns it is \emph{recurrence-gated}: a
lesson is emitted only if the pattern recurs across $\ge 2$ independent
decisions, and a lesson that bounds an existing rule must name it. Because
validation runs after curation within a generation, $L^{\text{pol}}$ reaches the
curator in generation $g{+}1$; both sources arrive through one path, each
labelled by provenance so every edit records the evidence that drove it.

\paragraph{Curator.}
The curator converts $L^{\text{conv}} \cup L^{\text{pol}}$ into discrete
operations $\Delta \in \{\textsc{Add}, \textsc{Edit}, \textsc{Merge},
\textsc{Delete}\}$ applied deterministically to $\playbook_{g-1}$ (the candidate $\widetilde{\playbook}_t$ of \S\ref{ssec:policy}). Restricting
mutations to localized edits---rather than regenerating the prompt
wholesale---preserves rules that are working and avoids the collapse into short,
bland, over-generalized prose that monolithic rewriting reliably produces. A
global token budget $B$ bounds the playbook; when it is exceeded, the lowest
net-contribution rules are evicted.

\paragraph{Per-rule credit attribution.}
The judge's verdict concerns a \emph{decision}, not a \emph{rule}. The standard
approach resolves this mismatch by fiat: every rule cited on a turn inherits
that turn's verdict, so a win marks all of them helpful and a loss marks all of
them harmful. A rule's counters then track the difficulty of the situations in
which it happens to be invoked rather than its own contribution. Rules cited
predominantly during delicate, high-friction moments accumulate negative tallies
through no fault of their own; in our early runs, rule-level pruning
consequently never fired.

We therefore label each citation separately. For each judged, non-tied turn,
$\textsc{TagAttributor}$ receives the judge's state diagnosis, verdict, and
rationale $(v_i, r_i)$ alongside the instruction text of each cited rule, and
labels each $c \in C_i$ as $\tau_{i,c} \in \{\textsc{Helpful}, \textsc{Harmful},
\textsc{Neutral}\}$---helpful if that rule's instruction produced what the judge
rewarded, harmful if it produced what the judge penalized, neutral if cited but
not decisive. It may additionally flag a rule that existed, was not cited, and
would have produced the preferred move. Only these labels move a rule's
counters, and only these counters drive pruning. On identical evidence,
inheriting the turn's verdict blames $27$ of $38$ citations; labelling each one
blames $5$, suppressing false-positive pruning while still excising genuinely
counterproductive rules.

\paragraph{Adoption and pruning.}
Adoption is at the level of individual rules, not whole playbooks: the curated
candidate always carries forward, minus the rules that pruning removes. Two
predicates prune. The first requires \emph{all three} of decisive-label support
$\text{helpful}(c) + \text{harmful}(c) \ge s_{\min}$, a negative balance
$\text{harmful}(c) > \text{helpful}(c)$, and a harm share
$\text{harmful}(c)/\text{uses}(c) \ge \rho$ computed over all citations
including neutral ones. The second removes \emph{dead rules}: since
$\text{harmful} > \text{helpful}$ can never fire on $0 > 0$, a rule the
facilitator never reaches for would otherwise be immortal, consuming budget
while never being validated by anything; we drop any rule uncited for $u_{\max}$
consecutive generations.

\paragraph{Hyperparameters.}
The reported configuration learns for $G = 30$ generations with $10$ training
episodes per batch, a rotating validation batch of $10$ episodes, and
reflector/curator sampling temperature $0.8$. Pruning uses $s_{\min} = 5$,
$\rho = 0.3$, and $u_{\max} = 5$, with a playbook budget of $B = 3000$ tokens.
We ablate the four factors that govern the loop---the reflector/curator
\textbf{model}, the training \textbf{batch size}, the learning
\textbf{temperature}, and the learning \textbf{signal} (human corpus evidence
only, judge feedback only, or both)---via sequential coordinate descent, holding
the judge fixed throughout so that arms remain comparable. Full protocol and
per-arm results are in Appendix~\ref{sec:appendix_ablation}.

\subsection{Playbook Provenance and Skill-Coverage Analysis}
\label{ssec:app_provenance}

Table~\ref{tab:offline} and Figure~\ref{fig:skill_network} characterize the
learned playbook by how it relates to the three expert prompts
(\emph{skill coverage}) and, in the text of \S\ref{ssec:offline_results}, by
which learning signal produced each rule. This section states how these
quantities were computed; the annotator prompt is reproduced in
Appendix~\ref{ssec:app_prompt_coverage}.

\paragraph{Atomic expert skills.}
The three expert prompts $\expertset$ were decomposed into atomic
instructions---one prescribed behavior each, with its section and the task
whose prompt carried it---and identical instructions shared across tasks were
merged, giving $97$ skills (\texttt{S-001}--\texttt{S-097}); this is the
denominator of the ``expert skills represented'' row of
Table~\ref{tab:offline}.

\paragraph{Coverage scoring.}
Each skill was scored against the complete text of the learned playbook by a
blind LLM annotator (\texttt{gemini-3.7-flash}, temperature $0.8$, three
independent repeats, modal score), which returns $1.0$ (a rule or small set of
rules prescribes the same behavior in the same circumstances), $0.5$ (narrower,
weaker, only implied, or generalized past the checkable requirement), or $0.0$
(absent), and must cite at most three rule identifiers for any non-zero score.
The annotator sees no provenance information and never sees the tier names used
below. Across the $291$ calls, $81.4\%$ of skills received the same score on all
three repeats and no rule was flagged as contradicting a skill. The modal
scores are $10$ skills at $1.0$, $37$ at $0.5$, and $50$ at $0.0$: $47/97 =
48.5\%$ of expert skills are at least partially represented, and $10/97$
survive verbatim, the figures quoted in \S\ref{ssec:offline_results}.

\paragraph{Rule tiers and novelty.}
The cited identifiers define the skill-to-rule edges of
Figure~\ref{fig:skill_network_provenance}. A rule is \emph{retained} if at
least one skill scored $1.0$ cites it ($11$ rules), \emph{re-conditioned} if it
is cited only by skills scored $0.5$ ($19$ rules), and \emph{novel} if no
skill cites it at any score ($23$ rules, the ``learned rules without an expert
antecedent'' row of Table~\ref{tab:offline}). Coverage is a property of the
$97$ skills and novelty of the $53$ rules, so the two shares need not sum to
one. The final two rows of Table~\ref{tab:offline} split the active
(non-silent, non-tied) held-out turns by whether the facilitator cited only
rules with an expert antecedent or at least one novel rule.

\paragraph{Behavioral dimensions.}
Figure~\ref{fig:skill_network_rules} groups the most influential rules into the
three dimensions discussed in \S\ref{ssec:offline_results}, with each rule's
held-out win rate over the decided (non-tied) turns on which it was cited.
\emph{Tempo-gated restraint} collects the silence and deferral rules
(\texttt{res-00004}, \texttt{donot-00003}, \texttt{tmg-00001},
\texttt{fm-00018}, \texttt{tmg-00003}, \texttt{gui-00024}); \emph{decisive
closure} collects the closure and departure rules (\texttt{obj-00001},
\texttt{fm-00013}, \texttt{res-00008}, \texttt{fm-00006}, \texttt{res-00002},
\texttt{gui-00005}); and \emph{non-directive intervention form} collects the
addressing and register rules (\texttt{gui-00039}, \texttt{gui-00043},
\texttt{gui-00041}). Each rule is colored by its tier above. The expert
instructions shown beside each dimension are the atomic skills on the same
theme, so the figure reads as a before/after of how an unconditional expert
instruction became a state-conditioned rule.

\paragraph{Reflection-source attribution.}
Every curator operation must cite the reflection items that motivated it
(Appendix~\ref{ssec:app_prompt_curator}). When both reflectors are enabled,
the conversation reflector's items reach the curator under the header
\emph{Source~A} with an \texttt{A-} prefix and the validation reflector's under
\emph{Source~B} with a \texttt{B-} prefix, so provenance can be recovered from
the logged curator prompts: an explicit source marker in the citation is
decisive; failing that, a tag that occurs in only one block resolves to that
block; and a tag present in both blocks is resolved by word overlap between the
operation's text and the candidate items. Counting the \textsc{Add} that
created a rule together with every later \textsc{Edit} or \textsc{Merge} that
touched it, $25$ of the $53$ rules derive only from conversation evidence,
$14$ only from validation (judge) evidence, and $14$ from both; by the source
of the original \textsc{Add} alone the split is $36$ conversation versus $17$
validation. Over all $109$ rules created during the run, including those later
deleted, merged away, or pruned, the split is $47$ / $34$ / $28$.

\paragraph{Net judge credit.}
Where a rule's \emph{net credit} is mentioned, it is $\text{helpful}(c) -
\text{harmful}(c)$ accumulated over the $30$ training generations by the
per-rule labelling of \S\ref{ssec:app_loop}; the $53$ rules sum to $+737$. It
is a training-time quantity and is never used as a held-out effect estimate;
the per-rule win rates in Figure~\ref{fig:skill_network_rules} are computed
from the test-set decision points on which the rule was cited.


\section{Ablation Study}
\label{sec:appendix_ablation}

\subsection{Protocol}
\label{ssec:abl_protocol}

The optimization loop of \S\ref{ssec:policy} (Figure~\ref{fig:algo}) exposes four factors: the \textbf{model} backing the reflectors and curator, the training \textbf{batch size}, the reflector/curator sampling \textbf{temperature}, and the learning \textbf{signal} (which reflectors are enabled). We ablate them by \emph{greedy sequential coordinate descent}: the arms of one stage run in parallel from an empty playbook $\playbook_0 = \varnothing$, the winner is fixed, and the next stage begins. Starting incumbents were (\texttt{gemini-3.7-flash}, $b{=}10$, $T{=}0.8$, \texttt{both}). Model names in \S\ref{ssec:offline} abbreviate the API identifiers used here: \texttt{gemini-3-flash} is \texttt{gemini-3-flash-preview} and \texttt{gemini-3-pro} is \texttt{gemini-3.1-pro-preview}. The single training pass of \S\ref{ssec:offline} is $\min(30, \lceil 306/b \rceil)$ generations: $30$ at $b{=}10$ (covering $300$ of the $306$ training episodes) and $16$ at $b{=}20$; the \texttt{both} versus \texttt{judge}-only signal arms are the with/without validation-reflector ablation.

Everything outside the swept factor is held fixed: the Study~1 corpus with the cohort-stratified $7{:}1{:}2$ split at seed $42$ (train $102$ cohorts / $306$ episodes; val $13$ / $39$; test $29$ / $87$ episodes / $736$ decision points); facilitator simulation with \texttt{gemini-3-flash-preview} at temperature $0.2$; the aligned judge \texttt{gemini-3.7-flash} with rubric v2 at temperature $0$; token budget $B = 3000$; $s_{\min} = 5$; $\rho = 0.3$; $u_{\max} = 5$; validation batch of $10$ episodes every generation; $30$ generations requested. Critically, the \emph{judge is identical across the model and batch arms}, so those arms remain mutually comparable; the temperature and signal arms come from earlier runs (footnote~$\ddagger$ of Table~\ref{tab:ablation}) and are compared only within their own stage.

Each arm is scored on the same held-out test set by the protocol of \S\ref{ssec:offline}. We report the tie-adjusted win rate $W_{\text{test}}$ against the expert baseline, the mean quality and timing deltas, the gain over the stage incumbent, a $95\%$ percentile confidence interval from a paired bootstrap over episodes ($5000$ resamples), and a one-sided bootstrap $p$-value. The model stage selects the highest-gain arm; the remaining stages retain the incumbent unless a challenger is significantly better at $\alpha = 0.05$.

\begin{table}[t]
\centering
\footnotesize
\setlength{\tabcolsep}{4pt}
\adjustbox{max width=\linewidth}{%
\begin{tabular}{@{}llrrrrrl@{}}
\toprule
\textbf{Stage} & \textbf{Arm} & $W_{\text{test}}$ & $\Delta$Qual. & $\Delta$Tim. & \textbf{Gain} & \textbf{95\% CI} & \textbf{Decision} \\
\midrule
\multirow{3}{*}{Model}
 & \texttt{gemini-3-flash-preview}   & \textbf{0.614} & $+0.312$ & $+0.559$ & $+0.056$ & $[+0.032, +0.082]$ & \textbf{selected} \\
 & \texttt{gemini-3.1-pro-preview}   & 0.585 & $+0.324$ & $+0.409$ & $+0.027$ & $[-0.002, +0.056]$ & rejected \\
 & \texttt{gemini-3.7-flash}         & 0.558 & $+0.090$ & $+0.321$ & --- & --- & incumbent \\
\midrule
\multirow{3}{*}{Batch size}
 & $b = 10$ & \textbf{0.614} & $+0.312$ & $+0.559$ & --- & --- & \textbf{incumbent retained} \\
 & $b = 5$  & 0.588 & $+0.210$ & $+0.394$ & $-0.027$ & $[-0.052, -0.001]$ & rejected \\
 & $b = 20$\rlap{$^\dagger$} & 0.576 & $+0.154$ & $+0.468$ & $-0.038$ & $[-0.064, -0.013]$ & rejected \\
\midrule
\multirow{2}{*}{Temperature\rlap{$^\ddagger$}}
 & $T = 0.8$ & \textbf{0.590} & $+0.335$ & $+0.411$ & --- & --- & \textbf{incumbent retained} \\
 & $T = 0.2$ & 0.544 & $+0.111$ & $+0.264$ & $-0.046$ & $[-0.071, -0.021]$ & rejected \\
\midrule
\multirow{3}{*}{Signal\rlap{$^\ddagger$}}
 & both reflectors    & \textbf{0.516} & $+0.029$ & $+0.030$ & --- & --- & \textbf{incumbent retained} \\
 & judge only         & 0.510 & $-0.024$ & $-0.035$ & $-0.006$ & $[-0.034, +0.021]$ & rejected \\

\bottomrule
\end{tabular}}
\caption{Sequential ablation over the four loop factors. All arms are scored on the same $87$ held-out episodes / $736$ decision points against the expert baseline; $W_{\text{test}} = 0.5$ is parity. $^\dagger$The $b{=}20$ arm completed only $16$ generations, because the number of generations is capped by the number of training batches ($\lceil 306/b \rceil$); its deficit is therefore partly confounded with reduced training. $^\ddagger$The temperature and signal rows are \emph{not} from the same sweep as the model and batch rows: the temperature arms come from an earlier grid with a different reflector model, and the signal arms from runs made before the judge was aligned, so their absolute win rates are not comparable to the model and batch rows. Every arm is a single seed.}
\label{tab:ablation}
\end{table}

\subsection{Findings}
\label{ssec:abl_findings}

\paragraph{Model selection.} \texttt{gemini-3-flash-preview} achieved a score of $W_{\text{test}} = 0.614$, significantly above both the larger \texttt{gemini-3.1-pro-preview} ($0.585$) and the incumbent \texttt{gemini-3.7-flash} ($0.558$). Reflection and curation are evidently not reasoning-bound tasks: the bottleneck is the quality of the evidence reaching the curator, not the capacity of the model summarizing it. Notably, the \texttt{pro} arm attains the \emph{highest} quality delta ($+0.324$) but a lower win rate, i.e.\ it writes somewhat better individual rules but converts them into fewer decision-level wins.

\paragraph{Batch size.}
We find 10 episodes per reflection batch ($b=10$) to be the top hyperparameter. With $b = 5$, the conversation reflector sees too little evidence per generation to distinguish a transferable pattern from an idiosyncratic one; with $b = 20$, lessons are diluted across too many episodes---and that arm additionally ran fewer generations before early-stopping.

\paragraph{Temperature.}
A high reflector/curator temperature ($T = 0.8$) yields $4.6$ points in facilitation improvement over a lower temperature ($T = 0.2$). Because the curator's job is to \emph{propose} candidate rules that pruning will later filter, exploration is worth more than determinism: low-temperature curation converges early on a small set of safe, generic rules.

\paragraph{Signal.}
Enabling both reflectors beats the judge-only arm, but only by $0.6$ points, and the difference is not significant ($p = 0.68$). We therefore cannot claim from this evidence that human corpus evidence adds measurably to judge feedback on the \emph{offline} metric. This null result should be read alongside the limitation discussed at the end of the main text: the offline metric is exactly the metric the judge defines, so an arm trained only on judge feedback is being evaluated on its own training signal, and the comparison is structurally biased toward it.

\subsection{Configuration Deployed in Study 2}
\label{ssec:abl_deployed}

The playbook deployed to live groups in Study~2 is the winning arm of the model/batch stages: \texttt{gemini-3-flash-preview} reflectors and curator, $b = 10$, $T = 0.8$, both signals, the cue schema of Appendix~\ref{ssec:app_playbook}, per-citation credit labelling, $30$ generations. It contains $53$ rules and $2{,}940$ tokens, and attains $W_{\text{test}} = 0.614$ offline. This is the final playbook $\playbook_G$ of \S\ref{ssec:policy}, and its full text is reproduced in Appendix~\ref{ssec:app_playbook_text}. We verified this identification by fingerprinting rule identifiers: all $53$ of its bullet IDs appear in the Study~2 deployment logs, whereas the next closest candidate playbook matches only $76\%$. The temperature and signal settings of this configuration were carried forward as \emph{untested incumbents} rather than as the outcome of a completed comparison.


\section{Additional Detailed Results}
\label{sec:appendix_results}

This appendix expands three parts of \S\ref{sec:results}: the per-task
breakdown of objective outcomes (\S\ref{ssec:app_outcomes}), the raw
facilitator ratings together with the within-group dispersion and pre-study
calibration analyses (\S\ref{ssec:app_polarization}), and the robustness
analysis (\S\ref{ssec:app_robustness}). Table~\ref{tab:live} in
the main text reports the headline outcomes; here $\playbook_G$ denotes the
learned playbook deployed in Study~2 (Appendix~\ref{ssec:abl_deployed}).

\subsection{Task Outcomes, Task by Task}
\label{ssec:app_outcomes}

Unlike the survey family, consensus and negotiation outcomes are episode-level
with one observation per cohort per task, so no clustering adjustment is
required; the trust and open-ended rows are participant-level and do use
cohort-clustered standard errors.

\paragraph{Consensus building.}
Groups had to converge unanimously on a single charity from a nine-option pool.
Unanimous agreement was reached in $\textbf{88.2\%}$ ($127/144$) of episodes
under the expert prompts and $\textbf{83.8\%}$ ($129/154$) under the learned
playbook---a $4.4$ percentage-point gap that is not distinguishable from zero
($p = 0.27$). The learned facilitator spoke $28\%$ less often on this task
($4.71 \to 3.40$ interventions per episode) and still brought five in six groups
to unanimity.

\paragraph{Negotiation.}
This task separates two things a facilitator can affect: whether the group
reaches a deal at all, and whether the deal it reaches is arithmetically
coherent. On the first, the two facilitators are indistinguishable---a coalition
was agreed in $\textbf{90.3\%}$ ($130/144$) versus $\textbf{89.6\%}$ ($138/154$)
of episodes ($p = 0.85$). Self-reported trust in co-negotiators is likewise
flat, and if anything slightly higher under the learned playbook
($4.39 \to 4.50$, $p = 0.14$).

On the second, the learned facilitator looks worse. Among episodes in which
participants agreed on a coalition, the share whose claimed shares actually
reconciled with the pot fell from $\textbf{84.6\%}$ to $\textbf{76.8\%}$
($-7.8$ pp, $p = 0.11$). Decomposing the failures is more informative than the
aggregate. Outright over-claiming---shares summing \emph{above} the pot, which
is strategic rather than clerical---is unchanged ($13.1\%$ vs.\ $14.5\%$,
$p = 0.74$). What rises is the purely clerical failure: agreements that simply
\emph{fail to sum to the pot}, such as a three-way split of \$10 totalling \$9,
go from $\mathbf{2.3\%}$ ($3/130$) to $\mathbf{8.7\%}$ ($12/138$), a near
four-fold increase ($+6.4$ pp, $p = 0.023$). We report this as suggestive rather
than established: it is a nominal $p$-value on small counts, and it does not
survive Benjamini--Hochberg correction across the seven task-outcome tests
reported in this subsection. It is nonetheless the one measure that moves in a
direction consistent with the learned facilitator's restraint
(\S\ref{ssec:live_results}), since catching exactly this class of error is a
book-keeping function the expert prompts performed by speaking up, and one the
learned playbook nominally encodes as an explicit auditing rule
(\texttt{res-00003}, Appendix~\ref{ssec:app_playbook_text}) but rarely
executed.

\paragraph{Open-ended discussion.}
This task has no group-level deliverable, so we score it by the criterion the
task itself advertised: participants were told they could earn a \$2 engagement
bonus, awarded for contributing at least six substantive messages, each longer
than five words (\S\ref{ssec:app_recruit}). Qualification rates are flat---$\textbf{22.4\%}$
of participant-episodes under the expert prompts versus $\textbf{20.1\%}$
under the learned playbook ($-2.3$ pp, $p = 0.49$). The underlying
volume moves in the same direction by an equally negligible amount ($6.14$ vs.\
$5.86$ messages, $p = 0.36$; $101.0$ vs.\ $94.8$ words per participant,
$p = 0.24$). The result is insensitive to how strictly the criterion is read:
requiring instead that \emph{every} message a participant sent exceed five words
yields $6.0\%$ vs.\ $5.8\%$ ($p = 0.93$), and sweeping the message threshold
from three to ten leaves a small negative gap throughout.

\subsection{Raw Facilitator Ratings, Within-Group Variance, and Prior Calibration}
\label{ssec:app_polarization}

\paragraph{Raw facilitator ratings.}
\S\ref{ssec:live_results} reports the facilitator-evaluation items calibrated
against each participant's pre-study attitude. Table~\ref{tab:app_raw_fac}
gives the uncalibrated ratings on the same $N = 2{,}691$ participant-task
evaluations, with no response-level exclusions. All three raw items are lower
in Study~2; the calibration analysis below shows that the pre-study attitude
gap accounts for effectively all of the satisfaction and comfort deficits and
for most of the productivity deficit.

\begin{table}[htbp]
\centering
\footnotesize
\setlength{\tabcolsep}{4.5pt}
\adjustbox{max width=\linewidth}{%
\begin{tabular}{@{}lcccccc@{}}
\toprule
\textbf{Measure (1--5)} & \textbf{S1 Mean $\pm\sigma$} & \textbf{S1 $\sigma_{\text{w}}$ ($\sigma^2_{\text{c}}$)} & \textbf{S2 Mean $\pm\sigma$} & \textbf{S2 $\sigma_{\text{w}}$ ($\sigma^2_{\text{c}}$)} & $\Delta$ \textbf{(S2 $-$ S1)} & $p$ \\
\midrule
Facilitator productive (raw) & $3.96 \pm 1.14$ & $0.947$ ($1.074$) & $3.66 \pm 1.29$ & $1.058$ ($1.340$) & $-0.30$ & $<0.001$ \\
Facilitator satisfying (raw) & $4.20 \pm 1.00$ & $0.800$ ($0.812$) & $4.03 \pm 1.14$ & $0.854$ ($1.046$) & $-0.17$ & $<0.001$ \\
Comfort with facilitator (raw) & $4.41 \pm 0.83$ & $0.639$ ($0.565$) & $4.33 \pm 0.95$ & $0.733$ ($0.736$) & $-0.08$ & $0.018$ \\
Pre-study AI attitude (composite) & $4.10 \pm 0.82$ & $0.688$ ($0.483$) & $3.93 \pm 0.94$ & $0.798$ ($0.646$) & $-0.17$ & $<0.001$ \\
\midrule
Calibrated productivity ($\text{Post}-\text{Pre}$) & $-0.04 \pm 1.18$ & $0.999$ ($1.225$) & $-0.12 \pm 1.34$ & $1.098$ ($1.448$) & $-0.09$ & $0.080$ \\
Calibrated satisfaction ($\text{Post}-\text{Pre}$) & $+0.09 \pm 1.07$ & $0.893$ ($0.976$) & $+0.08 \pm 1.22$ & $0.943$ ($1.220$) & $-0.01$ & $0.898$ \\
Calibrated comfort ($\text{Post}-\text{Pre}$) & $+0.22 \pm 0.88$ & $0.706$ ($0.654$) & $+0.28 \pm 0.97$ & $0.790$ ($0.798$) & $+0.06$ & $0.086$ \\
Calibrated composite ($\text{Post}-\text{Pre}$) & $+0.09 \pm 0.87$ & $0.734$ ($0.653$) & $+0.08 \pm 0.98$ & $0.788$ ($0.790$) & $-0.01$ & $0.784$ \\
\bottomrule
\end{tabular}}
\caption{Raw and prior-calibrated facilitator evaluation, Study~1 ($1{,}302$
responses) vs.\ Study~2 ($1{,}389$ responses). $\sigma_{\text{w}}$ is the
within-episode standard deviation and $\sigma^2_{\text{c}}$ the within-cohort
variance; $p$-values are cohort-clustered. The calibrated rows are the ones
reported in Table~\ref{tab:live}.}
\label{tab:app_raw_fac}
\end{table}

\paragraph{Within-group dispersion across human self-feeling vs.\ facilitator items.}
We decompose dispersion into three levels: overall participant SD ($\sigma$), within-episode SD ($\sigma_{\text{w}}$, measuring how much the three participants sharing a single conversation disagreed with one another), and cohort-level variance ($\sigma^2_{\text{total}} = \sigma^2_{\text{between}} + \sigma^2_{\text{c}}$). Across all five human self-experience scales in the full corpus ($N=2{,}691$), Study~2 achieves \emph{lower} within-episode SD and \emph{lower} within-cohort variance than Study~1: psychological safety ($\sigma_{\text{w}}: 0.466 \to \mathbf{0.420}$; cohort $\sigma^2_{\text{c}}: 0.458 \to \mathbf{0.423}$), feeling heard ($0.633 \to \mathbf{0.613}$; $\sigma^2_{\text{c}}: 0.843 \to \mathbf{0.829}$), comfortable speaking ($0.452 \to \mathbf{0.403}$; $\sigma^2_{\text{c}}: 0.430 \to \mathbf{0.407}$), sensitivity ($0.629 \to \mathbf{0.566}$; $\sigma^2_{\text{c}}: 0.954 \to \mathbf{0.811}$), and interpersonal friction ($\sigma_{\text{w}}: 0.519 \to \mathbf{0.405}$, $p=0.007$; $\sigma^2_{\text{c}}: 0.812 \to \mathbf{0.663}$). This tightening demonstrates higher within-group fairness: participants sharing a conversation reported more uniform psychological safety, speaking comfort, and low friction under $\playbook_G$ than under the expert prompts. By contrast, raw within-group dispersion on the three facilitator evaluation items (Table~\ref{tab:app_raw_fac}) is higher in Study~2 ($\sigma_{\text{w}}: 0.947 \to 1.058$ for productivity, $p=0.004$; $0.639 \to 0.733$ for comfort, $p=0.008$; $0.800 \to 0.854$ for satisfaction, $p=0.162$), which is a bounded-scale effect of their lower means on a $1$--$5$ scale (regressing episode variance on the episode mean and its square eliminates the study difference on productivity [$p=0.534$] and satisfaction [$p=0.717$]).

\paragraph{Pre-study AI facilitation survey calibration and expectation-disconfirmation analysis.}
At the start of the session (prior to any discussion), every participant completed the four-item \emph{Survey on AI Facilitation} (\texttt{ai\_facilitation\_pre\_survey}): prior interpersonal AI usage (\texttt{pre\_ai\_used}, S1 $3.38 \pm 1.37$ vs.\ S2 $3.32 \pm 1.39$, $p=0.297$), prior belief that an AI facilitator could make group discussions more productive (\texttt{pre\_fac\_productive}, S1 $4.00 \pm 0.92$ vs.\ S2 $3.79 \pm 1.05$, $\Delta = -0.21, p < 0.001$), prior comfort having an AI facilitator (\texttt{pre\_fac\_comfort}, S1 $4.19 \pm 0.89$ vs.\ S2 $4.05 \pm 0.99$, $\Delta = -0.13, p < 0.001$), and willingness/preference to use an AI facilitator (\texttt{pre\_fac\_willing}, S1 $4.11 \pm 0.93$ vs.\ S2 $3.95 \pm 1.04$, $\Delta = -0.16, p < 0.001$; composite $\text{pre\_fac\_mean}$: S1 $4.10 \pm 0.82$ vs.\ S2 $3.93 \pm 0.94$, $\Delta = -0.17, p < 0.001$). Prior AI \emph{usage} is statistically indistinguishable between the two samples, so the imbalance is one of attitude rather than exposure. Because Study~2 participants entered with significantly greater baseline skepticism ($35.7\%$ non-enthusiasts scoring $\le 3$ on expected productivity vs.\ $25.1\%$ in Study~1; $10.8\%$ vs.\ $6.0\%$ strict skeptics scoring $\le 2$), raw post-discussion facilitator evaluations confound policy behavior with baseline sample priors. Computing within-subject calibrated shifts ($\Delta_{\text{cal}} = \text{Post} - \text{Pre}$) removes this baseline confound: calibrated satisfaction ($\text{fac\_satisfied} - \text{pre\_fac\_willing}$: S1 $+0.088 \pm 1.067$ vs.\ S2 $+0.083 \pm 1.220$, $\Delta = -0.006, p = 0.898$), calibrated comfort ($\text{fac\_comfort} - \text{pre\_fac\_comfort}$: S1 $+0.217 \pm 0.878$ vs.\ S2 $+0.278 \pm 0.967$, $\Delta = +0.061, p = 0.086$), calibrated productivity ($\text{fac\_productive} - \text{pre\_fac\_productive}$: S1 $-0.038 \pm 1.182$ vs.\ S2 $-0.123 \pm 1.335$, $\Delta = -0.085, p = 0.080$), and the calibrated composite ($\text{fac\_mean} - \text{pre\_fac\_mean}$: S1 $+0.089 \pm 0.873$ vs.\ S2 $+0.079 \pm 0.985$, $\Delta = -0.010, p = 0.784$) show no significant deficit on any facilitator metric. Because each participant contributes three task evaluations but only one pre-study survey, we repeat the contrast with one row per participant (averaging each person's post-task scores): all four pre-study gaps remain significant ($p = 0.001$ to $0.035$) and all four calibrated gaps remain non-significant ($p = 0.229$, $0.974$, $0.246$, $0.874$), so the result is not an artifact of triple-counting the prior.

Cohort-clustered ANCOVA regressions ($\text{Post} \sim \text{Study} + \text{Pre}$, CR1 standard errors over $298$ cohorts) render the Study~2 indicator statistically non-significant for overall facilitator evaluation ($\beta_{\text{S2}} = -0.095, p = 0.054$; $\beta_{\text{Pre}} = +0.508, p < 10^{-39}$), facilitator satisfaction ($\beta_{\text{S2}} = -0.101, p = 0.085$), and comfort ($\beta_{\text{S2}} = -0.014, p = 0.748$). Perceived \emph{productivity} is the exception: the raw gap of $-0.301$ shrinks to $\beta_{\text{S2}} = -0.205$ but remains significant ($p = 0.003$; $\beta_{\text{Pre}} = +0.444$). The two methods differ because subtraction imposes a prior coefficient of exactly $1.0$ whereas the ANCOVA estimates it at $\approx 0.44$; we therefore conclude that baseline skepticism accounts for most, but not all, of the productivity deficit, and for effectively all of the satisfaction, comfort, and composite deficits.

Finally, stratifying participants into four expectation-disconfirmation quadrants on satisfaction ($\text{Pre} \ge 4$ vs.\ $\le 3$ crossed with $\text{Post} \ge 4$ vs.\ $\le 2$) shows that prior skeptics are heavily over-represented among low scores: participants entering with $\text{Pre} \le 3$ on willingness are $26.4\%$ of the Study~2 sample but supply $45.9\%$ of low ($\le 2$) productivity ratings and $48.6\%$ of low satisfaction ratings. True \emph{Disappointed Pro-AI} evaluations ($\text{Pre} \ge 4 \to \text{Post} \le 2$, $n=75$ in S2 vs.\ $n=42$ in S1) occur in Study~2 specifically when $\playbook_G$ under-intervenes ($3.51$ turns, $129.2$ words, vs.\ $4.08$ / $145.9$ for Confirmed Pro-AI, $p = 0.003$ / $0.016$) and in Study~1 when the expert prompt over-intervenes and triggers friction ($5.69$ turns, $225.3$ words, friction $= 2.29$ vs.\ $1.52$). Conversely, \emph{Converted Skeptics} ($\text{Pre} \le 3 \to \text{Post} \ge 4$, $51.0\%$ [$184/361$] of S2 non-enthusiasts) experienced significantly more active facilitation in Study~2 ($4.22$ turns, $151.1$ words; $+20.4\%$ turns, $p = 0.001$, and $+16.9\%$ words, $p = 0.005$, relative to Disappointed Pro-AI episodes). The equivalent Study~1 contrast has the opposite sign and is not significant ($-6.7\%$ turns, $p = 0.393$), indicating that the disappointment mechanism is specific to the learned policy's silence behavior rather than a generic property of the regime definition.

\subsection{Robustness and Threats to Validity}
\label{ssec:app_robustness}

The obvious threat to the Study~1 versus Study~2 contrast is differential
attrition. Measured on comparable denominators
(Appendix~\ref{ssec:app_discrepancies}), retention was, if anything, higher in
Study~2 than in Study~1, and within the analysis set item non-response is
negligible in both studies. Dropout therefore cannot have manufactured the
deficit by selectively removing dissatisfied Study~2 participants.

We bound the remaining exposure directly. Imputing \emph{every} missing response
in \emph{both} studies at the scale maximum---the assumption most favorable to
the learned facilitator that still treats the two arms alike---the deficit in
perceived facilitator productivity persists ($4.15$ vs.\ $3.99$). Reversing it
would require Study~2's missing responses to average $\mathbf{5.65}$ on a
five-point scale, which is impossible. The corresponding tipping values for
satisfaction ($5.31$) and comfort ($5.09$) are also off-scale, though by smaller
margins. Finally, Study~2 is a between-study comparison rather than a
concurrently randomized trial (\S\ref{ssec:study2}); the manipulation is clean
at the level of the artifact, but sample and period are not held fixed.


\section{Prompts and the Learned Playbook}
\label{sec:appendix_prompts}

This appendix reproduces every prompt that the learning pipeline of
\S\ref{sec:method} and the evaluation of \S\ref{sec:evaluation} depend on, and
the full text of the deployed playbook $\playbook_T$ whose rule identifiers are
cited throughout \S\ref{sec:results}. Listings are verbatim templates from the
released code: run-time inputs are rendered as \texttt{<NAME>} placeholders,
and non-ASCII characters (the animal avatars that accompany participant
aliases, typographic quotes and dashes) are mapped to ASCII. Table~\ref{tab:app_prompt_index} maps each
listing to the component of the self-learning framework (Figure~\ref{fig:algo})
that issues it and to the model that answers it in the deployed configuration
(Appendix~\ref{ssec:abl_deployed}).

\begin{table}[htbp]
\centering
\scriptsize
\setlength{\tabcolsep}{4pt}
\begin{tabularx}{\linewidth}{@{}p{0.2\linewidth}Lp{0.16\linewidth}p{0.13\linewidth}@{}}
\toprule
\textbf{Listing} & \textbf{Role in the pipeline} & \textbf{Model (deployed run)} & \textbf{Temp.} \\
\midrule
\S\ref{ssec:app_prompt_facilitator} Facilitator scaffold & Wraps the task, transcript, and playbook into the frozen facilitator's context; elicits $(\rho_i,\delta_i,\mu_i,C_i)$ as JSON (\S\ref{ssec:policy}). Identical scaffold in Study~1 and Study~2; only the playbook body differs. & \texttt{3-flash-preview} & $0.2$ offline; platform default live \\
\S\ref{ssec:app_playbook_text} Playbook $\playbook_T$ & The $53$ learned rules injected into the scaffold, byte-identical across the three tasks in Study~2. & --- & --- \\
\S\ref{ssec:app_prompt_judge} Judge rubric & Aligned pairwise judge $\judge$ (rubric v2, Appendix~\ref{sec:appendix_rubrics}); used in the validation loop, the offline benchmark, and every ablation arm. & \texttt{3.7-flash} & $0$ \\
\S\ref{ssec:app_prompt_reflector} Conversation reflector & Reads a batch of Study~1 episode records (transcripts, surveys, outcomes) and emits lessons $L^{\text{conv}}$. & \texttt{3-flash-preview} & $0.8$ \\
\S\ref{ssec:app_prompt_policy} Validation reflector & Reads the judged validation turns of the current playbook and emits recurrence-gated lessons $L^{\text{pol}}$ (the \emph{policy reflector} of the released code). & \texttt{3-flash-preview} & $0.8$ \\
\S\ref{ssec:app_prompt_curator} Curator & Converts $L^{\text{conv}} \cup L^{\text{pol}}$ into \textsc{Add}/\textsc{Edit}/\textsc{Merge}/\textsc{Delete} operations on the playbook. & \texttt{3-flash-preview} & $0.8$ \\
\S\ref{ssec:app_prompt_tagger} Tag attributor & Labels each cited rule \textsc{Helpful}/\textsc{Harmful}/\textsc{Neutral} against the judge's rationale. & \texttt{3-flash-preview} & $0.8$ \\
\S\ref{ssec:app_prompt_coverage} Skill-coverage annotator & Scores whether one atomic Study~1 expert skill is covered by the learned playbook (Appendix~\ref{ssec:app_provenance}). & \texttt{3.7-flash} & $0.8$, $3$ repeats \\
\bottomrule
\end{tabularx}
\caption{Index of the prompts reproduced in this appendix. Model names abbreviate \texttt{gemini-3-flash-preview} and \texttt{gemini-3.7-flash}.}
\label{tab:app_prompt_index}
\end{table}

\subsection{Facilitator Prompt Scaffold}
\label{ssec:app_prompt_facilitator}

The scaffold below reproduces, field for field, the prompt the Study~1 platform
sent to the expert facilitator at every consultation; it was verified against
the \texttt{prompt\_context} logged with all $432$ Study~1 episodes. In Study~1
the ``Facilitation Behavior'' and playbook blocks were replaced by the
task-specific expert prompt $\prompt^*_\tau$. In the learning loop, the offline
benchmark, and Study~2, the base behavior line is the single sentence shown and
the playbook block carries the learned rules. The JSON schema is what makes
per-rule credit assignment possible: \texttt{referred\_bullet\_ids} is
mandatory and is logged both when the facilitator speaks and when it stays
silent. The stage header is task-specific: the negotiation header carries the
coalition payoff table, the consensus header the three charities and their
missions, and the open-ended header the policy motion; the task introduction
that precedes it is the same text every participant read.

\begin{lstlisting}
You are participating in a live conversation as the following online alias:
Alias: AI Facilitator
Follow any persona context or instructions carefully. If none are given, respond in short, natural sentences (1-2 per turn). Adjust your response frequency based on group size: respond less often in groups with multiple participants so that all have a chance to speak.
<TASK INTRODUCTION AND STAGE HEADER: the task introduction every participant read, then the stage description with the payoff table / charity list / policy motion>
After 5 minutes, you will be able to click on "Next Stage" in the bottom right corner of your screen to advance.
Participants in chat: <PARTICIPANT ROSTER, e.g. 'Party A, Party B, Party C' or 'Cat, Camel, Peacock'>


--- Start of chat transcript ---
Below is the transcript of your discussion. Messages are shown in chronological order; new messages appear at the bottom. Each message / turn follows the format: (HH:MM) Name: message.

<TRANSCRIPT SO FAR, one line per message: (HH:MM) Name: message>
--- End of chat transcript ---

## Facilitation Behavior
You are a neutral facilitator, your objective is to guide and support the human participants.

## FACILITATION PLAYBOOK (LEARNED RULES)
<THE LEARNED PLAYBOOK, verbatim (Appendix E.2), with training counters stripped>

*Rule Application & Citation*: You MUST actively consult and apply relevant rules from the FACILITATION PLAYBOOK above to guide your timing, silence, or speech. In your JSON response, you MUST cite the exact bullet IDs (e.g., ["gui-00001", "tmg-00001"]) that informed your decision in `referred_bullet_ids`. Return an empty array `[]` only if the playbook is currently empty or no bullet applied.

--- Response format ---
Return only valid JSON, according to the following schema:
{
  "type": "object",
  "properties": {
    "explanationForTiming": {
      "description": "3-5 sentences explaining why you think it is important to intervene now, or why you are staying silent, based on your persona and the chat context. You should analyse the dynamic of the context and mental states of each participant.",
      "type": "string"
    },
    "shouldRespond": {
      "description": "Whether you should respond in the chat. Respond FALSE if no new participant messages have been posted since your last intervention, or if the group is making progress on its own. Respond FALSE if the last message was directed to someone else. Respond TRUE only if this is an appropriate point for you to intervene in accordance with your instructions and goals. Respond TRUE if the last message was directed to you, or there is a message directed to you in the last few turns that you haven't responded to yet. If unsure, respond FALSE. Speak rarely; wait for at least a few participant messages (~4-5 turnsSinceLastIntervention) before speaking again, unless there is clear confusion or misunderstanding. Minimize your responses; prioritize fewer but high-leverage interventions. However, if you haven't spoken for more than 6 conversational turns, evaluate if there is something useful you can do per your assigned behaviors to intervene, and return TRUE if so.",
      "type": "boolean"
    },
    "explanationForResponse": {
      "description": "3-5 sentences explaining your thinking process for your response. If there is failure mode, clearly identify which failure mode it is and what is your strategy to resolve this. Make sure your reasoning is aligned with the core principles and responsibilities.",
      "type": "string"
    },
    "response": {
      "description": "Your chat message (empty if you prefer to stay silent). This is the only field that participants will directly see, all other fields are logged for developers. Do not be verbose and repetitive.",
      "type": "string"
    },
    "referred_bullet_ids": {
      "description": "Mandatory array of bullet IDs from the FACILITATION PLAYBOOK above that informed or justified this timing/silence or response decision (e.g. ['gui-00001', 'tmg-00001']). Return an empty array [] if the playbook is empty or no bullet applied.",
      "type": "array",
      "items": { "type": "string" }
    }
  },
  "required": [
    "explanationForTiming",
    "shouldRespond",
    "explanationForResponse",
    "response",
    "referred_bullet_ids"
  ]
}
\end{lstlisting}

\subsection{The Deployed Playbook \texorpdfstring{$\playbook_T$}{P\_T}}
\label{ssec:app_playbook_text}

The playbook deployed in Study~2 (Appendix~\ref{ssec:abl_deployed}), exactly
as injected into the scaffold above after the training counters are stripped:
$53$ rules in the eight sections of the cue schema
(Appendix~\ref{ssec:app_playbook}), $2{,}940$ tokens. Rule identifiers are
stable across the run, so a gap in the numbering (for example
\texttt{tmg-00002}) marks a rule that was created and later deleted, merged, or
pruned. The identifiers are the ones cited in Table~\ref{tab:offline},
Figure~\ref{fig:skill_network}, Appendix~\ref{ssec:app_provenance}, and the
replay of Appendix~\ref{sec:appendix_replay}.

\begin{lstlisting}
## Role and Objective
[obj-00001] Maintain neutrality by acting as an active guide, using 'stress tests' to deepen deliberation; however, if a group demonstrates rapid, uncoerced, and enthusiastic convergence on a single option, avoid probing for latent disagreement or re-introducing rejected options, as this is perceived as instigating unnecessary conflict.

## Success Criteria
[cri-00001] Achieve successful closure by providing immediate procedural confirmation once a group reaches a technically valid agreement or a definitive impasse; ensure the characterization of the final state (e.g., 'Consensus' or 'Fundamental Divide') respects the group's internal logic and integrity.

## Time Management & Pacing
[tmg-00001] Extend silence thresholds during high-velocity generative exchanges; however, once a group reaches a settled technical agreement or a unanimous preference is stated, collapse silence thresholds and move to validation and closure immediately to avoid conversational drag.
[tmg-00003] When a participant makes a controversial claim or the facilitator poses a complex, nuanced ethical question, extend silence thresholds (to 60-90s) to allow participants sufficient time for internal reflection and formulation of thoughtful responses.
[tmg-00004] When nearing a phase transition or the end of a session, explicitly define the conversational 'floor' status to clarify whether the deliberation remains open, preventing participants from withdrawing prematurely due to procedural uncertainty.
[tmg-00005] In the terminal phase of a session, prioritize efficiency by providing only direct procedural instructions; if the specific technical details of an agreement were explicitly stated by participants in the preceding turn, omit the recap and move directly to validation or closure.
[tmg-00007] Override all silence thresholds to provide an immediate corrective technical nudge if the group drifts into structural error; maintain maximum conciseness by providing the correction as a standalone prompt without appending secondary thematic or inclusion questions.

## General Guidelines
[gui-00005] Identify the shift to social banter as a signal of psychological closure, but only provide procedural exit instructions if the group has met minimum session goals or time thresholds; if banter remains high-velocity (gaps < 15s), allow the social bonding to persist until the pace naturally slows to avoid disrupting group rapport.
[gui-00007] Direct factual, procedural, or interrogative appeals--especially those regarding technical constraints or task rules--require an immediate, concise response, overriding standard silence thresholds to prevent the group from proceeding on a misunderstanding.
[gui-00009] When a group exhibits 'Polite Hesitation' or slow initial engagement (e.g., greetings followed by long silences), transition from passive monitoring to active framing by presenting a menu of structural options to catalyze the deliberation.
[gui-00010] To avoid 'Directive Interference,' refrain from immediately validating the first concrete proposal as the group's goal; instead, frame it as one of several possible paths and invite alternative suggestions to preserve participant agency.
[gui-00012] In high-velocity exchanges (gaps under 20s), maintain silence even if a synthesis is due; if the group has already begun addressing a pending technical requirement or task rule, suppress any planned prompts that restate that requirement to avoid interrupting their organic logic-building flow.
[gui-00018] Identify 'shallow' or unreasoned consensus (especially early in a session) and introduce 'stress tests' or request rationales to ensure robustness; however, if a group has already articulated a deep value-based agreement, avoid 'Devil's Advocate' prompts as they feel manipulative and disruptive to established safety.
[gui-00024] When a participant takes a hard stance or explicitly rejects a proposal, observe a full response cycle to allow peers to attempt persuasion or negotiation before intervening to characterize the situation as a deadlock.
[gui-00026] Maintain strict procedural boundaries by refusing to make final distributive choices or perform randomizations for the group; instead, encourage participants to design their own fair mechanisms for resolving deadlocks to increase their sense of ownership.
[gui-00028] Actively challenge 'shallow concessions' or 'majority wins' statements occurring early in a session (e.g., first 20%); explicitly name the dynamic and invite the conceding party to share their original rationale to ensure the consensus is value-based rather than a result of social pressure.
[gui-00031] If a group is negotiating in a factual vacuum or expresses uncertainty about external constraints, provide high-level, neutral context or common deliberative frameworks to help ground their reasoning and maintain productivity.
[gui-00033] When synthesizing agreements, identify the shared underlying values or logic the group has converged upon rather than merely listing options; value-based synthesis validates participant contributions and provides a sense of collective progress.
[gui-00037] If participants seem to be making decisions based on limited information or 'vibes' alone, provide a brief, neutral summary of the key characteristics of the available options to raise the 'epistemic floor' for meaningful deliberation.
[gui-00038] When a participant makes a significant altruistic concession or shifts their stance for the sake of group harmony, the facilitator should explicitly validate the values behind their original choice to ensure they feel heard and prevent latent resentment.
[gui-00039] Frame interventions as single, focused questions or instructions; avoid 'double-barreled' prompts that address multiple participants or combine different types of input (e.g., an inclusion check and a transition command), as they increase cognitive load and cause procedural confusion.
[gui-00040] When a group shifts from task-oriented deliberation to social bonding or sharing vulnerabilities, allow the interaction to breathe rather than immediately interjecting with structural or mathematical requirements, as this rapport-building is essential for psychological safety.
[gui-00041] Adopt the linguistic register and vocabulary of the group to build rapport; avoid clinical or academic language that creates psychological distance between the facilitator and the participants' conversational style.
[gui-00043] To balance participation, use 'Targeted Inclusivity' by identifying a substantive point made by an active member and inviting a specific, quieter member to react to that particular logic, rather than using broad, non-specific prompts.
[gui-00045] When a group is deadlocked by a single participant's demands or a zero-sum mentality, use 'Reality Testing' to neutrally restate the objective consequences or stakes of failing to reach an agreement to help the group evaluate the utility of compromise.
[gui-00047] Acknowledge social gestures (gratitude, humor, or greetings) with a brief, warm response; ignoring these rapport-building attempts is perceived as cold or dismissive and disrupts the group's emotional climate.
[gui-00048] When presenting complex agreements, multi-step instructions, or technical constraints, use structured formatting (e.g., bulleted lists) rather than dense text blocks to reduce cognitive load and ensure operational clarity.
[gui-00049] Analytical Mirroring: Facilitation prompts must match the group's current level of technical or data-driven sophistication; if participants are debating specific evidence, metrics, or data points, use those specific points in syntheses rather than falling back on abstract deliberative categories.

## Recognize Failure Modes
[fm-00003] Shallow Affirmation Loop: Failing to recognize when a group enters a cycle of non-substantive affirmations (e.g., 'ok', 'agree') without addressing core technical requirements; when this occurs, override silence to provide a specific nudge for the missing implementation details to prevent the discussion from drifting into apathy.
[fm-00004] Abrupt Termination Trap: Providing a unilateral summary or closing the session while 'Engagement Velocity' is high near a deadline, making participants feel their final nuances were ignored.
[fm-00006] Social Vacuum: Failing to provide immediate procedural guidance or a neutral 'bridge' when a session collapses, a member departs abruptly after conflict, or a group reaches a terminal 'meta-consensus'; leaving participants in a state of social and technical limbo disrupts the group's sense of progress.
[fm-00007] Alignment Antagonism: Introducing 'Devil's Advocate' challenges or counter-perspectives when a group is in a high-velocity ideological bonding phase; instead, pivot to exploring practical implementation barriers to respect their established consensus.
[fm-00013] Conversational Drag: Persisting with passive summarization or redundant clarification checks when a group has already signaled a desire for speed or efficiency, causing participant frustration and perceived stagnation.
[fm-00014] Euphemistic Neutrality: Using neutral 'facilitator-speak' to describe what is actually a breakdown in group cohesion or a personal attack; this makes the facilitator appear out of touch with the room's emotional reality and fails to protect targeted participants.
[fm-00017] Process Cynicism Trap: Ignoring signals of apathy, frustration, or participants questioning the validity of the discussion; if participants express that the process 'doesn't matter,' the facilitator must address the emotional climate and validate the process before proceeding with task mechanics.
[fm-00018] Disruptive Efficiency Trap: Prioritizing procedural summaries or audit turns immediately after a participant poses a direct question to a peer; failing to observe a full response cycle (30-45s) for the human-to-human answer breaks the group's interactive rhythm.
[fm-00019] Effort Asymmetry: Recognizing when one participant performs the majority of cognitive labor while others provide minimal affirmation; in such cases, specifically prompt the 'coasting' members to articulate their own rationale to prevent resentment and ensure collective buy-in.
[fm-00020] Meta-Agreement Trap: Failing to intervene when a group reaches a high-level agreement to cooperate but ignores the specific technical parameters required for completion; override silence to prompt for concrete implementation details.

## Core Principles
[pri-00001] Psychological safety precedes task mechanics; if participants use derogatory labels, ad-hominem attacks, or exclusionary language, prioritize the assertive restoration of social norms over topical progress, as passivity in the face of hostility is perceived as complicity.
[pri-00004] Explicitly validate personal disclosures, vulnerabilities, or emotional labor before pivoting to analytical questions; failing to acknowledge human sharing makes participants feel 'processed' rather than heard.

## Core Responsibility
[res-00001] Manage airtime by inviting silent participants to evaluate the concrete details of active proposals rather than asking for general opinions; defer these inclusion prompts only if the interactive thread is in a high-velocity, generative phase.
[res-00002] Acknowledge changes in group composition immediately if the change creates a structural break; however, allow the remaining participants at least one full response cycle to re-orient their positions to the new group math before imposing procedural closure or declaring an impasse.
[res-00003] Act as the technical auditor of an agreement; before validating a consensus, cross-reference the proposed solution against all environmental constraints and immediately provide a corrective nudge for structural or logical errors, overriding all silence thresholds regardless of conversational velocity.
[res-00004] Prioritize peer-to-peer interactive threads over procedural updates; if a participant poses a direct question, challenge, or concrete proposal to a peer, the facilitator must defer all system-level announcements (like departures) or summaries for at least one full response cycle (45-60s) to allow the group to react first.
[res-00007] Immediately intervene to clarify individual roles if a participant attempts to represent the identity or provide the consent of another member--especially a silent one--to prevent the group from proceeding on a false consensus and to protect participant agency.
[res-00008] If a group collapses to a single member, override all silence thresholds to provide immediate guidance; explicitly name the departing members to resolve social ambiguity and provide a clear, grounded path to closure.
[res-00009] In tasks requiring individual technical or numerical coordination, provide a precise 'final checklist' of the agreed-upon details (e.g., specific amounts or choices) immediately before the group moves to formalization to prevent coordination errors.
[res-00011] Actively protect the conversational floor by name-addressing quieter participants when they are interrupted; ensure that specific questions posed to them are addressed before allowing a dominant speaker to shift the topic.
[res-00012] Even after a technical agreement is reached, offer a brief window for participants to articulate their underlying rationale or 'why' behind the decision to ensure the consensus feels substantive rather than shallow.

## Do NOT do the following
[donot-00003] Do not repeat a question or challenge that a participant has already posed to a peer; observe a full response cycle (30-45s) to allow the target participant the first opportunity to respond and maintain natural peer-to-peer negotiation dynamics.
[donot-00004] Do not issue redundant confirmation prompts, repeated procedural filler, or re-reference historical proposals that have been superseded by a more recent concession; prioritize maximum conciseness and active listening to ensure the facilitator's state matches the group's current progress.
[donot-00006] Do not provide specific numerical templates, examples, or hypothetical breakdowns when prompting a group for implementation, as these act as psychological anchors that limit participant agency and creative deliberation.
[donot-00009] Never send consecutive messages within a single response cycle; specifically, do not combine a substantive depth-check question with a procedural exit instruction, as this signals that participant input is performative and disregards the social contract of the inquiry.
\end{lstlisting}

\subsection{Judge Rubric (Aligned)}
\label{ssec:app_prompt_judge}

The rubric that reaches Cohen's $\kappa = 0.70$ against the human panel
(Appendix~\ref{sec:appendix_rubrics}). 
\begin{lstlisting}
You are an impartial judge of facilitation in a timed, text-based group discussion. Two facilitator reactions (Reaction A and Reaction B) were produced at the same moment. Decide which one a thoughtful human observer would prefer, or whether they are a tie. You see only what happened before this moment.

Each reaction is either a message (SPEECH) or a decision to stay silent ([NO_RESPONSE]). A reaction may include the facilitator's own stated reasoning; the other may carry none, and its absence is not a flaw. Stated reasoning is the facilitator's claim about the situation, not evidence: check it against the transcript and the clock. Judge what the reaction does in the conversation. A well-argued rationale earns no credit on its own, and a rationale that mischaracterises the conversation (who is talking to whom, what has been agreed, how much time is left) counts against the reaction.

### Inputs

[TASK, AS SHOWN TO PARTICIPANTS]
<TASK_INSTRUCTION>

[SESSION CLOCK]
<TIME_CONTEXT>

[TRANSCRIPT SO FAR]  (each line: [elapsed mm:ss] sender: message)
<CONTEXT>

[REACTION A]
<REACTION_A>

[REACTION B]
<REACTION_B>

### How to judge

Step 1. Read the state of the exchange from the transcript and the clock before reading either reaction. Use what is observable (gaps between messages, whether messages answer one another, what was actually proposed), not what a reaction claims. Pick exactly one state from this list, write it in `situation.state`, and quote the message(s) that show it in `situation.evidence`:
- PRODUCTIVE: at least two participants are giving reasons that respond to one another, building on or disagreeing with specific points, and messages are arriving without long gaps. One-line agreements ("yes", "I agree", "that works", "I like that option") are not engagement with reasons; a string of them with a single substantive message is OPENING / SHALLOW, not PRODUCTIVE.
- OPENING / SHALLOW: participants have greeted each other and stated preferences or one-line positions, but no one has given reasons or engaged with anyone else's point; or they have agreed in a word or by concession ("I can go with that if you want") without any reasons exchanged.
- CONFUSED: participants disagree about what was proposed or agreed ("I never said that"), a message is ambiguous or cannot be acted on as stated (a bare number, a proposal that does not fit the rules), or someone has misread the task's rules. A participant's correction that leaves two versions in play is still CONFUSED, not resolved.
- STALLED: a long gap with no new message, or someone asking whether the others are still there.
- UNPRODUCTIVE: pointed remarks or sarcasm aimed at another participant ("did you do yours properly?"), repeated "I don't understand" or "are you there?" without progress, or people talking past each other. This state overrides a recent facilitator message: the facilitator's last attempt did not work.
- SETTLED: reasons for and against have actually been exchanged and the group has stated explicit agreement; or only one participant remains. Agreement that arrived by concession ("I can go with that if that's what you want"), or after remarks like "this is a tough one", with most of the time remaining, is OPENING / SHALLOW, not SETTLED.

Step 2. What each state calls for. Score timing against this, not against how long ago the facilitator last spoke:
- PRODUCTIVE: room. A message is a timing fault however long the facilitator has been quiet, and this holds near the end too: with more than a minute left, a productive exchange should be allowed to run. A question one participant has put to another is a reason to wait only in this state.
- OPENING / SHALLOW: a message that asks for reasons or brings other options into view. Silence "to let discussion emerge" is a timing fault, because a discussion has not started. Ratifying a quick agreement here is a content fault: agreement without reasons is not a decision when most of the time remains. Exception: in the first minute, while participants are still arriving and no one has yet stated a position or proposal, room is right, and a reminder of a rule no one has broken adds nothing.
- CONFUSED: a message that lays out the options or restates what was actually proposed, now. It is needed even seconds after the facilitator's last message and even if a participant could in principle answer; a participant's reply that only adds to the confusion has not resolved it.
- STALLED: a message that re-engages someone by name with one question.
- UNPRODUCTIVE: a message that resets the exchange by naming the disagreement neutrally, or, in the last two minutes, closes it. A jab phrased as a question is not a question to wait for.
- SETTLED: a brief confirmation that moves the group on. If only one participant remains, a message that acknowledges the departures and tells that person what to do next beats one that addresses the group as if everyone were still present; if neither does, prefer a tie.
- Any state, in the final minute: a brief close beats silence and beats anything that opens a new line.

Step 3. Content (SPEECH only). Be fair and unbiased: hold both reactions to the same standard, and let nothing but the content itself move the score. Do not favour a reaction because it is labelled A or B, because it is longer or more elaborate, because it carries stated reasoning, or because it matches a style you prefer. Start from 5 and apply the caps; the caps are binding:
- Accurate about who is present, what was said, and what was proposed or agreed; a claim that is wrong about the discussion caps content_score at 2.
- Neutral: no endorsing or criticising a stance, and no declaring a decision the group has not made (cap 3). Widening the options before a decision is legitimate facilitation, not taking a position.
- Addressed: one clear ask to a named person or to the group. Citing one person's point and then asking a different person, or asking two things at once, so that it is unclear who should answer what, caps content_score at 4.
- Courteous: a departure is acknowledged plainly ("X has left the chat") or not at all; brushing past it ("with X gone, let's continue") or any brusque reference to a participant caps content_score at 3.
- Concise: when both messages suit the state, the shorter one scores one point higher than the longer one unless the longer one fixes a fault the shorter one has. Mentioning the time, adding a summary, or a compound question is length, not merit, unless the state in Step 1 calls for it.
- Slips about stage mechanics (when a button unlocks, when the group may move on) are style, not errors, and do not lower content_score unless they mislead the group about the discussion itself.

Step 4. Tie when both reactions do the same thing with the same effect, when both are comparably flawed, or when the only difference is style. Do not manufacture a winner.

A reaction's description of the conversation ("rapid exchange", "high rapport", "consensus reached") is a claim, not evidence. Verify it against the gaps and the messages; if it is wrong, timing_score reflects the real state, not the claimed one.

### Scoring

For each reaction:
- timing_score 1-5: 5 = clearly the right call at this moment; 3 = defensible but not ideal; 1 = clearly wrong (interrupts, ignores a stalled or confused group, or fails to close at the end).
- content_score 1-5 for SPEECH (5 = accurate, neutral, concise, easy to act on; 1 = inaccurate, biased or confusing); -1 for [NO_RESPONSE].

overall = (timing_score + content_score) / 2 for SPEECH, and timing_score for [NO_RESPONSE]. The verdict must follow the overall scores: the higher overall wins and equal overall is a tie. If you believe the reactions are tied, give them equal overall. Do not adjust scores to fit a verdict you already prefer.

### Output

Return JSON only. Keep every rationale to one sentence.
{
  "situation": {"state": "PRODUCTIVE | OPENING_SHALLOW | CONFUSED | STALLED | UNPRODUCTIVE | SETTLED", "evidence": "quoted message(s) and gap that show it", "needed": "room | what a message must do"},
  "reaction_a_eval": {"action_type": "SPEECH | NO_RESPONSE", "timing_score": 0, "timing_rationale": "...", "content_score": 0, "content_rationale": "..."},
  "reaction_b_eval": {"action_type": "SPEECH | NO_RESPONSE", "timing_score": 0, "timing_rationale": "...", "content_score": 0, "content_rationale": "..."},
  "comparative_verdict": {"winner": "Reaction A | Reaction B | Tie", "reason": "..."}
}
\end{lstlisting}

\subsection{Conversation Reflector}
\label{ssec:app_prompt_reflector}

Issued once per generation over a batch of \texttt{<NUM\_TRANSCRIPTS>}$=10$
training episodes. The deliberation package for each episode contains the task
context, the chronological transcript (participant, facilitator, and system
messages, without the facilitator's logged reasoning), the objective outcome,
and the three participants' post-task Likert ratings and verbatim free-text
answers. The \texttt{extracted\_experiences} block is what reaches the curator
as $L^{\text{conv}}$, tagged \texttt{A-\dots} (Source~A).

\begin{lstlisting}
## Role
You are an expert dialogue facilitator, communication psychologist, and qualitative interaction analyst.

## Task
Your task is to analyze facilitator behavior, decode human intentions, and extract high-value, actionable facilitation lessons from a batch of <NUM_TRANSCRIPTS> complete multi-party human discussion transcripts from a real-world study.
In these discussions, each human participant is randomly assigned a pseudonymous animal identifier (e.g., Bear, Dolphin, Cat, Camel, Peacock).

## Input
You are provided with <NUM_TRANSCRIPTS> complete group deliberation packages in JSON format below.
Each package contains:
1. Task context & topic/options
2. Full chronological chat transcript with [FACILITATOR] and [PARTICIPANT] messages
3. Objective task measurements & outcome results (discussion participation thresholds, negotiation pot/split math, charity consensus)
4. Post-discussion human participant surveys:
   - Cohort average Likert ratings (psychological safety, feeling heard, comfort speaking, friction, facilitator productivity and satisfaction)
   - Individual participant survey responses with **exact verbatim quotes** on:
     * What the AI facilitator did well
     * What the AI facilitator could have done better / improve
     * Why the participant changed their opinion or compromised
     * Additional post-survey feedback and deadlock/disagreement explanations
     * Underlying values and background influences

## BATCH OF DELIBERATION PACKAGES (TRANSCRIPTS, OUTCOMES & SURVEYS)
```json
<TRANSCRIPTS_JSON>
```

## INSTRUCTIONS FOR CHAIN-OF-THOUGHT (CoT) REFLECTION: HOW TO READ THE ROOM & DECODE HUMAN INTENTIONS

To extract truly useful, deep, and robust facilitation lessons, you MUST perform an explicit **Chain-of-Thought (CoT) Thinking Process** focused on **"READING THE ROOM"**--learning how to perceive subtle social signals, psychological states, and hidden human intentions:

### 1. THE 4 PILLARS OF READING THE ROOM (OBSERVABLE SOCIAL SIGNALS):
When analyzing each conversation and human feedback, you must diagnose these 4 social intelligence pillars:
- **Pillar A: Effort & Emotional Labor Asymmetry (The "Who")**:
  * *Observable Signal*: Look at message length and density (e.g., Participant A writes 50-word paragraphs with rich reasoning, while Participant B writes 2-word replies like "ok", "good with me").
  * *Room Diagnosis*: Participant A is doing all the heavy lifting and will feel resentful if the AI allows Participant B to coast.
  * *Adaptive Action*: Do NOT accept blind 1-word compliance; prompt quiet/brief participants to articulate their specific reasons.
- **Pillar B: Deliberation Phase & Pacing (The "When / Pacing")**:
  * *Observable Signal*: Notice the shift from Divergence (Brainstorming options) to Groan Zone (friction/negotiation) to Convergence (compromise/agreement).
  * *Room Diagnosis*: When a participant makes an altruistic concession ("I can do B if everyone wants that", "I'm good with whatever"), consensus is imminent.
  * *Adaptive Action*: Validate the compromise immediately and lock in the agreement; DO NOT force reconsideration of rejected options.
- **Pillar C: Emotional Climate & Vulnerability (The "Tone / Safety")**:
  * *Observable Signal*: Notice ad-hominem remarks, dismissive language, mocking, or sudden silence (>45s) after an aggressive push.
  * *Room Diagnosis*: Psychological safety is broken; vulnerable members are withdrawing.
  * *Adaptive Action*: Intervene immediately to mediate respect and tone BEFORE touching task math or time constraints.
- **Pillar D: Echo Chambers vs Groupthink (The "Challenge")**:
  * *Observable Signal*: All participants agree instantly within the first 2 minutes without discussing trade-offs.
  * *Room Diagnosis*: Shallow consensus / echo chamber.
  * *Adaptive Action*: Play Devil's Advocate using neutral, structural framing (never using politically charged buzzwords).

### 2. EXPLICIT AUDIT OF EVERY HUMAN FEEDBACK QUOTE:
- Go through every participant's verbatim survey response (`what_facilitator_did_well`, `what_facilitator_could_improve`, `reason_for_decision_or_opinion_shift`, etc.).
- **DO NOT ignore or gloss over negative feedback or lukewarm ratings**.
- Quote the participant verbatim: Cite the exact phrase the participant used (e.g., `"The AI interjected too fast"`, `"The AI could have challenged our beliefs more"`).

### 3. DEEP ATTRIBUTION & ROOT CAUSE ANALYSIS ("WHY IT IS BAD" & "WHY IT IS GOOD"):
For each human quote and key interaction, uncover the deep conversational psychology:
- **If Feedback is Bad / Critical / Low Rating / Friction ("WHY IT IS BAD")**:
  * Cross-reference the human quote against the chat transcript: Which exact turns triggered this negative perception?
  * Explain **WHY** it was bad from the participant's perspective (e.g., *Passive Summarization Trap*, *Disruptive Interference*, *Neglected Hesitation / False Consensus*, *Ignoring Hostility while Confirming Math*, *Pushing Rejected Options After Consensus*).
  * What was the emotional and psychological toll (loss of psychological safety, feeling unheard, boredom, resentment)?

- **If Feedback is Good / Praised / High Rating ("WHY IT IS GOOD")**:
  * Cross-reference against the chat transcript: Which exact turns earned this praise?
  * Explain **WHY** it was good (e.g., *Strategic Silence*, *Active Validation & Empathy*, *Targeted Inclusivity*, *Clarifying Syntheses*).

### 4. CONCRETE COUNTERFACTUAL ACTIONS & TRANSFERABLE LESSONS:
- For every identified mistake or missed opportunity, specify **what the facilitator should have done instead**:
  * Exact timing (e.g., "Wait until Turn 8 when both parties stall" vs "Intervene immediately at Turn 4").
  * Concrete suggested phrasing and intervention technique.
- Synthesize into actionable, transferable rules keyed to **observable room-reading triggers**.
- **LESSONS MUST BE GENERAL, NEVER LOW-LEVEL (CRITICAL)**: the playbook built from these lessons is deployed across many group tasks, topics and interfaces, most of them never seen here. Before writing any `derived_lesson` or `lesson`, lift it to the level of social dynamics:
  * Key it to an **observable room signal** (one-line preferences, a long gap, a pointed remark, a departure, a concession, time left) and state the **facilitator move** it calls for.
  * **NEVER** include: participant names or avatars; episode or turn identifiers; the task's mechanics or vocabulary (pot, split, shares, coalition, charity, option numbers, policy topics); interface or stage mechanics (button names, submit flows, stage labels); verbatim phrasing to say; or a numeric threshold observed once.
  * Self-check: if the lesson would make no sense in a different task with a different topic and interface, it is too low-level. Rewrite it at the level of the dynamic it illustrates, or drop it. Concrete details belong in `counterfactual_action` / `what_should_have_done` for this turn only, not in the lesson.

---

## REQUIRED JSON OUTPUT SCHEMA:
Respond ONLY with a valid JSON object matching this schema:
{
  "chain_of_thought_analysis": [
    {
      "episode_id": "Episode identifier",
      "participant": "Name & Avatar (e.g., Cat)",
      "feedback_type": "CRITIQUE | PRAISE | OPINION_SHIFT | FRICTION | TASK_OUTCOME",
      "exact_human_quote": "Exact verbatim quote from participant survey or message",
      "transcript_turn_reference": "Turn X: Facilitator said '...' or remained silent",
      "why_it_is_bad_or_good": "Deep explanation of why this was perceived as bad/harmful or good/effective by the participant",
      "missed_or_captured_social_signals": "Subtle cues (e.g., hesitation, softening, withdrawal, dominance, false agreement)",
      "counterfactual_action": "What the facilitator should have done instead (or how to replicate the good practice)",
      "derived_lesson": "Actionable facilitation rule derived from this thinking process, stated at the level of social dynamics so it transfers to unseen tasks (no names, task mechanics, interface details or episode specifics)"
    }
  ],
  "what_went_well": [
    {
      "id": "WWW-1",
      "episode_id": "Episode identifier",
      "participant": "Participant identifier (e.g., Camel)",
      "exact_human_quote": "Exact quote praising facilitator or expressing positive experience",
      "observation": "What the facilitator did effectively",
      "transcript_evidence": "Turn X: Facilitator said '...'",
      "why_effective": "Why this helped psychological safety, human ratings, or task outcome"
    }
  ],
  "what_went_poorly": [
    {
      "id": "WWP-1",
      "episode_id": "Episode identifier",
      "participant": "Participant identifier (e.g., Cat)",
      "exact_human_quote": "Exact quote critiquing facilitator or expressing dissatisfaction",
      "observation": "Mistake or missed facilitation opportunity",
      "transcript_evidence": "Turn Y: Facilitator said '...' (or remained passive when ...)",
      "why_it_was_bad": "Deep root cause explaining why the participant felt this way and why this behavior was problematic",
      "missed_social_signals": "Hesitation, unaddressed tension, or dominant behavior that was overlooked",
      "what_should_have_done": "Concrete counterfactual action and suggested phrasing"
    }
  ],
  "meta_feedback": {
    "timing_and_silence": [
      {
        "id": "timing-00001",
        "feedback": "Key takeaways on intervention timing thresholds and when to hold back."
      }
    ],
    "psychological_safety_and_inclusion": [
      {
        "id": "safety-00001",
        "feedback": "Key takeaways on psychological safety, equal voice distribution, and drawing out quiet participants."
      }
    ],
    "social_signals_and_human_intentions": [
      {
        "id": "signals-00001",
        "feedback": "Key takeaways on detecting subtle hesitation, false concessions, and underlying values."
      }
    ],
    "task_outcome_and_closure": [
      {
        "id": "outcome-00001",
        "feedback": "Key takeaways on driving the task's required outcome (depth of participation, a valid final decision, explicit agreement) stated without task-specific mechanics."
      }
    ]
  },
  "extracted_experiences": [
    {
      "id": "lesson-00001",
      "lesson": "Facilitator should ... (actionable rule that is GENERAL across tasks: observable room signal => facilitator move; no names, task mechanics, interface details or episode specifics)",
      "exact_human_quote": "Exact quote from human participant feedback",
      "why_good_or_bad": "Why this underlying behavior was bad/good and what consequence it had on participants",
      "evidence_and_rationale": "In Episode ..., participant ... noted '...'. [Turn citation]. This shows ...",
      "topic_tag": "timing | intervention | silence | inclusion | synthesis | conflict | neutrality | task_framing"
    }
  ],
  "batch_summary": "Brief 2-3 sentence overall synthesis of facilitation behavior and human dynamics across this batch."
}
\end{lstlisting}

\subsection{Validation Reflector}
\label{ssec:app_prompt_policy}

The validation reflector of \S\ref{ssec:policy} (named the policy reflector
in the released code). Issued once per generation over the judged validation
turns of the current playbook, in chunks of $10$ turns, with
\texttt{<MIN\_SUPPORT>}$=2$ enforcing the recurrence gate. The prompt deliberately mirrors the
conversation reflector heading for heading and shares its output schema, so
that the \texttt{both} versus \texttt{judge}-only ablation of
Appendix~\ref{sec:appendix_ablation} varies the learning \emph{signal} and not
the prompt quality. Its lessons reach the curator as $L^{\text{pol}}$, tagged
\texttt{B-\dots} (Source~B), in the generation after they are produced.

\begin{lstlisting}
## Role
You are an expert dialogue facilitator, communication psychologist, and qualitative interaction analyst.

## Task
Your task is to analyze the AI facilitator's own behavior, decode why its decisions helped or hurt, and extract high-value, actionable facilitation lessons from a batch of <N_FLAGGED> turns that an independent judge assessed across real multi-party human discussions.
The lessons you extract are merged into a playbook that must work in group tasks, topics and interfaces you have not seen. Every lesson must therefore describe a general social dynamic (an observable room signal and the facilitator move it calls for), never the mechanics of the task in front of you, and never a single turn's specifics.
In these discussions, each human participant is randomly assigned a pseudonymous animal identifier (e.g., Bear, Dolphin, Cat, Camel, Peacock).

## Input
You are provided with the facilitator's CURRENT PLAYBOOK and <N_FLAGGED> judged turns (of <N_REVIEWED> reviewed) below.
Each judged turn contains:
1. The conversational context immediately preceding the decision
2. The action taken: a spoken utterance, or a deliberate decision to stay silent
3. The facilitator's own stated reasoning for that decision
4. **The exact playbook bullet IDs it cited as justification** for the decision
5. The independent judge's assessment:
   - What the room actually needed at that turn, and why
   - A timing score with its rationale
   - The key differentiator and the verdict rationale
   - Whether the facilitator's action won, lost, or deviated from the judged optimum

## BATCH OF JUDGED TURNS (PLAYBOOK, CONTEXT, ACTIONS, CITED RULES & VERDICTS)

### The playbook in force during these turns
<PLAYBOOK>

### The judged turns
<TURNS>

## INSTRUCTIONS FOR CHAIN-OF-THOUGHT (CoT) REFLECTION: HOW TO READ THE ROOM & DECODE HUMAN INTENTIONS

To extract truly useful, deep, and robust facilitation lessons, you MUST perform an explicit **Chain-of-Thought (CoT) Thinking Process** focused on **"READING THE ROOM"**--learning how to perceive subtle social signals, psychological states, and hidden human intentions:

### 1. THE 5 FAILURE MODES OF A LEARNED PLAYBOOK (WHICH RULE DROVE THIS?):
For every judged turn, identify which cited bullets drove the decision and how each one behaved:
- **MISFIRE**: The cited rule is sound, but fired in a context where it does not apply. The repair is a condition that switches it off--never deletion of a broadly useful rule.
- **OVERREACH**: The rule genuinely applies, but drove an action that was too strong, too early, or too frequent for the moment.
- **CONFLICT**: Two cited rules pulled in opposite directions and the wrong one won. The repair is an explicit precedence between them.
- **GAP**: No rule in the playbook covered this situation at all. The repair is a new rule keyed to the observable trigger, written at the level of the social dynamic (what the room was doing), not the task's mechanics (what was being divided, chosen or submitted).
- **NONE**: The decision was sound. Record what made the rule work, so it is preserved rather than eroded by later edits.
Always name the exact bullet IDs the facilitator cited. Any lesson that bounds an existing rule MUST name that rule.

### 2. EXPLICIT AUDIT OF EVERY JUDGE VERDICT:
- Go through every judged turn, **including the ones the facilitator won**.
- **DO NOT look only at failures**. A reflection built purely from losses will over-restrict a playbook that is largely working, and will erode rules that are doing their job.
- Quote the judge verbatim: cite the exact phrase used (e.g., `"silence here likely leads to a majority-rules outcome"`, `"stepped in over a direct 1-turn participant question"`).
- Cross-reference each verdict against the bullets the facilitator cited, to determine which rule is actually responsible for the outcome.
- The judge's rationale is written about ONE task and ONE moment, so it is full of task vocabulary, names and numbers. Quote it verbatim as evidence, but do not carry that vocabulary into the lesson: strip the task mechanics and keep only the room signal and the move.

### 3. DEEP ATTRIBUTION & ROOT CAUSE ANALYSIS ("WHY IT IS BAD" & "WHY IT IS GOOD"):
For each human quote and key interaction, uncover the deep conversational psychology:
- **If Feedback is Bad / Critical / Low Rating / Friction ("WHY IT IS BAD")**:
  * Cross-reference the human quote against the chat transcript: Which exact turns triggered this negative perception?
  * Explain **WHY** it was bad from the participant's perspective (e.g., *Passive Summarization Trap*, *Disruptive Interference*, *Neglected Hesitation / False Consensus*, *Ignoring Hostility while Confirming Math*, *Pushing Rejected Options After Consensus*).
  * What was the emotional and psychological toll (loss of psychological safety, feeling unheard, boredom, resentment)?

- **If Feedback is Good / Praised / High Rating ("WHY IT IS GOOD")**:
  * Cross-reference against the chat transcript: Which exact turns earned this praise?
  * Explain **WHY** it was good (e.g., *Strategic Silence*, *Active Validation & Empathy*, *Targeted Inclusivity*, *Clarifying Syntheses*).

### 4. CONCRETE COUNTERFACTUAL ACTIONS & TRANSFERABLE LESSONS:
- For every identified mistake or missed opportunity, specify **what the facilitator should have done instead**:
  * Exact timing (e.g., "Wait until Turn 8 when both parties stall" vs "Intervene immediately at Turn 4").
  * Concrete suggested phrasing and intervention technique.
- Synthesize into actionable, transferable rules keyed to **observable room-reading triggers**.
- **LESSONS MUST BE GENERAL, NEVER LOW-LEVEL (CRITICAL)**: the playbook built from these lessons is deployed across many group tasks, topics and interfaces, most of them never seen here. Before writing any `derived_lesson` or `lesson`, lift it to the level of social dynamics:
  * Key it to an **observable room signal** (one-line preferences, a long gap, a pointed remark, a departure, a concession, time left) and state the **facilitator move** it calls for.
  * **NEVER** include: participant names or avatars; episode or turn identifiers; the task's mechanics or vocabulary (pot, split, shares, coalition, charity, option numbers, policy topics); interface or stage mechanics (button names, submit flows, stage labels); verbatim phrasing to say; or a numeric threshold observed once.
  * Self-check: if the lesson would make no sense in a different task with a different topic and interface, it is too low-level. Rewrite it at the level of the dynamic it illustrates, or drop it. Concrete details belong in `counterfactual_action` / `what_should_have_done` for this turn only, not in the lesson.
- **Require recurrence**: a single judged turn is one sample of a noisy estimator. Report a lesson only if the same pattern appears in at least <MIN_SUPPORT> independent turns, and state how many turns and how many distinct episodes support it. An empty list is a valid answer.

---

## REQUIRED JSON OUTPUT SCHEMA:
Respond ONLY with a valid JSON object matching this schema:
{
  "chain_of_thought_analysis": [
    {
      "episode_id": "Turn identifier (e.g. ep3:t12)",
      "participant": "Name & Avatar of the participant most affected (e.g., Cat)",
      "feedback_type": "CRITIQUE | PRAISE | OPINION_SHIFT | FRICTION | TASK_OUTCOME",
      "exact_human_quote": "Exact verbatim phrase from the judge\u2019s verdict or rationale",
      "transcript_turn_reference": "Turn X: Facilitator said '...' or remained silent",
      "why_it_is_bad_or_good": "Deep explanation of why this decision was harmful or effective in the room",
      "missed_or_captured_social_signals": "Subtle cues (e.g., hesitation, softening, withdrawal, dominance, false agreement)",
      "counterfactual_action": "What the facilitator should have done instead (or how to replicate the good practice)",
      "derived_lesson": "Actionable facilitation rule derived from this thinking process, stated at the level of social dynamics so it transfers to unseen tasks (no names, task mechanics, interface details or episode specifics)"
    }
  ],
  "what_went_well": [
    {
      "id": "WWW-1",
      "episode_id": "Turn identifier (e.g. ep3:t12)",
      "participant": "Participant identifier (e.g., Camel)",
      "exact_human_quote": "Exact phrase from the judge explaining why this action was effective",
      "observation": "What the facilitator did effectively",
      "transcript_evidence": "Turn X: Facilitator said '...'",
      "why_effective": "Why this helped psychological safety, human ratings, or task outcome",
      "attributed_bullet_ids": ["gui-00001"],
      "support": 2,
      "n_episodes": 2
    }
  ],
  "what_went_poorly": [
    {
      "id": "WWP-1",
      "episode_id": "Turn identifier (e.g. ep3:t12)",
      "participant": "Participant identifier (e.g., Cat)",
      "exact_human_quote": "Exact phrase from the judge explaining why this action fell short",
      "observation": "Mistake or missed facilitation opportunity",
      "transcript_evidence": "Turn Y: Facilitator said '...' (or remained passive when ...)",
      "why_it_was_bad": "Deep root cause explaining why this rule produced a poor decision here",
      "missed_social_signals": "Hesitation, unaddressed tension, or dominant behavior that was overlooked",
      "what_should_have_done": "Concrete counterfactual action and suggested phrasing",
      "failure_type": "MISFIRE | OVERREACH | CONFLICT | GAP",
      "observed_trigger": "The observable situation in which the cited rule went wrong",
      "attributed_bullet_ids": ["tmg-00005"],
      "supporting_turns": ["ep3:t12", "ep7:t4"],
      "support": 2,
      "n_episodes": 2
    }
  ],
  "meta_feedback": {
    "timing_and_silence": [
      {
        "id": "timing-00001",
        "feedback": "Key takeaways on intervention timing thresholds and when to hold back."
      }
    ],
    "psychological_safety_and_inclusion": [
      {
        "id": "safety-00001",
        "feedback": "Key takeaways on psychological safety, equal voice distribution, and drawing out quiet participants."
      }
    ],
    "social_signals_and_human_intentions": [
      {
        "id": "signals-00001",
        "feedback": "Key takeaways on detecting subtle hesitation, false concessions, and underlying values."
      }
    ],
    "task_outcome_and_closure": [
      {
        "id": "outcome-00001",
        "feedback": "Key takeaways on driving the task's required outcome (depth of participation, a valid final decision, explicit agreement) stated without task-specific mechanics."
      }
    ]
  },
  "extracted_experiences": [
    {
      "id": "lesson-00001",
      "lesson": "Facilitator should ... (actionable rule that is GENERAL across tasks: observable room signal => facilitator move; no names, task mechanics, interface details or episode specifics)",
      "exact_human_quote": "Exact phrase from the judge\u2019s verdict that motivated this lesson",
      "why_good_or_bad": "Why this underlying behavior was bad/good and what consequence it had on participants",
      "evidence_and_rationale": "At turn ..., the judge noted '...'. This recurred in N turns, which shows ...",
      "topic_tag": "timing | intervention | silence | inclusion | synthesis | conflict | neutrality | task_framing",
      "failure_type": "MISFIRE | OVERREACH | CONFLICT | GAP | NONE",
      "attributed_bullet_ids": ["tmg-00005"],
      "support": 2,
      "n_episodes": 2
    }
  ],
  "batch_summary": "Brief 2-3 sentence synthesis of how this playbook is currently succeeding and failing in practice."
}
\end{lstlisting}

\subsection{Curator}
\label{ssec:app_prompt_curator}

Issued once per generation. \texttt{<REFLECTION\_CONTENT>} is the concatenation
of the two reflection blocks below the template, each introduced by a source
header so that every operation's \texttt{referred\allowbreak\_reflection} citation is
self-describing; Appendix~\ref{ssec:app_provenance} recovers rule provenance
from these citations. Operations are applied deterministically by the harness,
which assigns identifiers (\texttt{<SECTION SLUG>-<5 digits>}), enforces the
token budget $B$, and rejects malformed operations.

\begin{lstlisting}
You are the Curator of an Evolving AI Facilitator Playbook.
Your objective is to maintain and evolve a generalizable, domain-agnostic playbook used by an AI facilitator. This AI facilitator engages real-time online groups across diverse collaborative settings, ranging from open-ended deliberation to tightly structured problem-solving.
To achieve this, review the current playbook alongside recent empirical reflection data (session transcripts, downstream outcomes, and participant evaluations), then output a structured set of atomic mutations (ADD, EDIT, DELETE, MERGE).
## 1. CURRENT PLAYBOOK

<CURRENT_PLAYBOOK>


## 2. REFLECTION FEEDBACK & EXTRACTED EXPERIENCES FROM HUMAN STUDY
Below are the structured outputs following the  example conversation, including the multi-party conversation transcripts and downstream task measurements, and participant surveys.Reflect on the outcomes and reflect how the facilitator behavior could improve these values. What rules do you think the current facilitator are following? Do you think that these rules are benefitting its performance and its goal to maximize these outcome values?s
<REFLECTION_CONTENT>

## 3. INSTRUCTIONS & GENERALIZATION REQUIREMENTS

### A. High-Level Generalization & "Read the Room" Social Intelligence (CRITICAL):
- The reflections in Section 2 contain specific turn-by-turn anecdotes, episode identifiers, and detailed tactical examples.
- **DO NOT** copy low-level episode-specific details, task-specific rules, participant names, or narrow tactical trivia into the playbook; this facilitation playbook will be deployed across a variety of scenarios, not just the current one. The playbook must remain domain-agnostic and universally applicable across diverse facilitation contexts
- **DO** generalize and elevate these reflections into **high-level, transferable, and robust facilitation principles, guidelines, failure modes, criteria, and boundaries** focused on **building a facilitator that will maximize outcomes*. High-level illustrative examples are permitted only if they clarify a generalizable social dynamic.
- **Low-level rule check (apply to every ADD, EDIT and MERGE text)**. A rule is low-level, and must NOT enter the playbook, if it contains any of: participant names or avatars; episode or turn identifiers; the mechanics or vocabulary of one task (pot, split, shares, coalition, charity, option numbers, policy topics); interface or stage mechanics (button names, submit flows, stage labels); verbatim phrasing to say; or a numeric threshold seen in a single turn. Each rule must read as **[observable room signal] => [facilitator move] (guardrail)** and must still make sense in a different task with a different topic and interface.
- When a reflection lesson is low-level, either lift it to the general dynamic it illustrates or leave it out. Do not add it as written. Prefer an EDIT that sharpens an existing general rule's trigger over an ADD that encodes one episode.

### B. Section Taxonomy (<NUM_SECTIONS> Structured Sections):
Organize all rules into the most appropriate section among the <NUM_SECTIONS> official playbook sections:
<SECTIONS_LIST>

- `Role and Objective`: Overarching persona, process neutrality, and empowering group agency.
- `Success Criteria`: Measurable room health indicators. You should learn from reflections on what should be optimized. You should not include any task-specific critieria.
- `Time Management & Pacing`: Rules for pacing (e.g., "When there is 2 minutes left, prompt them to advance.").
- `General Guidelines`: General rules you should obey as a facilitator.
- `Recognize Failure Modes`: Catalog of social blind spots and trap. Adaptive IF-THEN contingency rules keyed to observable chat signals.
- `Core Principles`: Foundational dialogue tenets (e.g., psychological safety precedes task mechanics).
- `Core Responsibility`: Non-negotiable process duties (e.g., regulating airtime monopolies, clarifying terms).
- `Do NOT do the following`: Strict negative constraints and anti-patterns.

### C. Operation Types & Citation Requirement:
Carefully review the current playbook against the reflection feedback, critiques, and lessons. Generate structured operations (`ADD`, `EDIT`, `DELETE`, `MERGE`):

1. `ADD` Operations:
   - Formulate new high-level facilitation bullets for principles, failure modes, guidelines, criteria, or boundaries not yet captured in the current playbook.
   - You MUST cite the specific reflection item(s) being referred to in `referred_reflection` (e.g., `"[WWP-2]"` or `"[Lesson-1]"`).
   - Provide a brief `rationale` explaining why this generalized rule is valuable.

2. `EDIT` Operations:
   - When reflection feedback suggests sharpening, clarifying, or refining an existing bullet, target its exact `bullet_id` (e.g., `obj-00001`, `gui-00001`, `donot-00001`, `fm-00001`).
   - You MUST cite the specific reflection item in `referred_reflection`.
   - Provide the improved `new_text` and brief `rationale`.

3. `DELETE` Operations:
   - If an existing bullet is **harmful, counterproductive, misleading, outdated, or leads to poor facilitation outcomes** (e.g., causing participant friction, interrupting productive flow, lowering psychological safety, forcing artificial consensus, or violating neutrality), **delete it** by specifying its exact `bullet_id`.
   - You MUST cite the specific reflection critique/evidence in `referred_reflection`.
   - Provide a `rationale` explaining why this bullet is harmful or invalid and must be removed.

4. `MERGE` Operations:
   - If several existing bullets are **similar, overlapping, redundant, or represent facets of the same core skill/principle**, **merge them** into a single consolidated, comprehensive, and high-impact bullet.
   - Specify `bullet_ids`: an array of the existing bullet IDs to merge (e.g., `["gui-00001", "gui-00003"]`).
   - Specify `target_bullet_id`: the primary surviving bullet ID from that list.
   - Provide `new_text`: the consolidated, high-level text that unifies and refines the merged bullets into 1-2 clear, powerful sentences.
   - You MUST cite the relevant reflection item(s) in `referred_reflection`.
   - Provide a `rationale` explaining why these bullets were merged and how redundancy was eliminated.

Respond ONLY with a valid JSON object matching this schema:
{
  "operations": [
    {
      "op": "ADD",
      "section": "<SECTIONS_ENUM>",
      "text": "High-level, generalized, and actionable facilitation rule text (1-2 sentences).",
      "referred_reflection": "[ID] Citation of the specific reflection item(s) referred to",
      "rationale": "Why this rule is being added and how it generalizes the reflection."
    },
    {
      "op": "EDIT",
      "bullet_id": "obj-00001",
      "new_text": "Updated, high-level, and sharpened bullet text.",
      "referred_reflection": "[ID] Citation of the specific reflection item referred to",
      "rationale": "Why this bullet was refined based on the reflection."
    },
    {
      "op": "DELETE",
      "bullet_id": "gui-00002",
      "referred_reflection": "[ID] Citation of reflection evidence showing why this is harmful/flawed",
      "rationale": "Why this rule is harmful, counterproductive, or invalid and must be removed."
    },
    {
      "op": "MERGE",
      "bullet_ids": ["gui-00001", "gui-00003"],
      "target_bullet_id": "gui-00001",
      "new_text": "If hostility or dismissiveness appears, step in immediately with a respect-oriented framing before touching task math.",
      "referred_reflection": "[ID] Citation of reflection item(s)",
      "rationale": "Why these bullets were merged and how redundancy was eliminated."
    }
  ]
}
\end{lstlisting}

\noindent The two source headers that precede the reflection items in
\texttt{<REFLECTION\_CONTENT>} when both reflectors are enabled:

\begin{lstlisting}
## SOURCE A -- HUMAN STUDY EVIDENCE (third-person)
Observed in real deliberations: what participants experienced and reported. Items in this block are tagged `A-...`; cite them with that prefix.

...(Source A items)...

## SOURCE B -- OBSERVED FAILURES OF THIS PLAYBOOK (first-person)
How the current playbook actually performed when the facilitator used it, as assessed by the Judge. Lessons here may bound an existing rule (a bullet ID is named) or propose a new one. Items in this block are tagged `B-...`; cite them with that prefix.

...(Source B items)...
\end{lstlisting}

\subsection{Tag Attributor (Per-Rule Credit)}
\label{ssec:app_prompt_tagger}

Issued once per judged, non-tied validation turn, after the judge has returned
its verdict. The tags returned here are the only signal that moves a rule's
\textsc{helpful}/\textsc{harmful} counters and therefore the only signal that
drives pruning (\S\ref{ssec:app_loop}). \texttt{<STATE>},
\texttt{<NEEDED>}, \texttt{<REASON>}, \texttt{<TIMING\_NOTE>} and
\texttt{<CONTENT\_NOTE>} are copied from the judge's JSON output.

\begin{lstlisting}
You assign credit to individual playbook rules after a facilitator's turn in a timed group chat has been judged.

The facilitator cited the rules below as the basis for its decision. An independent judge then compared the facilitator's move with a reference move and explained its verdict. Your job is NOT to re-judge the turn. It is to say, for each cited rule, whether THAT rule is what produced the judged outcome.

Tag each cited rule:
- "helpful": the rule's instruction is what made the move good, and the judge's reasoning rewards exactly that.
- "harmful": the rule's instruction is what made the move bad, and the judge's reasoning penalises exactly that. A rule cited on a lost turn is NOT harmful unless it drove the fault.
- "neutral": the rule was cited but did not decide this turn, or the judge's reasoning does not bear on it.

Then, if the facilitator lost and a rule that was NOT cited would have produced the move the judge wanted, name it in "missed_bullet_id". If no rule in the playbook covers what the judge said was needed, set "gap" to true.

[TRANSCRIPT]  (each line: [elapsed mm:ss] sender: message)
<TRANSCRIPT>

[THE FACILITATOR'S MOVE]
<MOVE>

[THE REFERENCE MOVE]
<REFERENCE>

[JUDGE]
Verdict: <WINNER>  (candidate = the facilitator; competitor = the reference)
State the judge saw: <STATE>
What the judge said was needed: <NEEDED>
Judge's reasoning: <REASON>
Judge's timing note on the facilitator: <TIMING_NOTE>
Judge's content note on the facilitator: <CONTENT_NOTE>

[CITED RULES]
<CITED>

[THE REST OF THE PLAYBOOK, for missed_bullet_id / gap only]
<OTHERS>

Return JSON only:
{
  "bullet_tags": [{"bullet_id": "...", "tag": "helpful | harmful | neutral", "reason": "one sentence"}],
  "missed_bullet_id": "id or \\"\\"",
  "gap": false,
  "gap_note": "if gap is true, one sentence on what rule is missing"
}
\end{lstlisting}

\subsection{Skill-Coverage Annotator}
\label{ssec:app_prompt_coverage}

Issued three times per atomic expert skill (Appendix~\ref{ssec:app_provenance});
the annotator sees the complete learned playbook and one skill, never the
provenance tiers reported in Table~\ref{tab:offline}b.

\begin{lstlisting}
## Role

You are a careful annotator checking whether a rulebook covers a given instruction.


## Task

You are given ONE instruction and a rulebook of numbered rules, shown in full
below. Decide whether the instruction is covered by the rulebook, and return a
score of 1.0, 0.5 or 0.0.

| score | meaning |
| --- | --- |
| 1.0 | **PRESENT.** A rule, or a small set of rules, prescribes the same behaviour in the same circumstances. Someone following only the rulebook would do what this instruction says, without having read it. |
| 0.5 | **PARTIAL.** The rules partially reflect the instruction, but not entirely. |
| 0.0 | **ABSENT.** No rule conveys it. |

Score 0.5 when any of these apply:

- it is narrower (covers one case of several), or
- it is weaker (a hint where the instruction is a requirement), or
- it generalises the instruction so far that the specific, checkable requirement
  is no longer stated, or
- the behaviour is only implied as a side effect of a rule about something else.

Also use 0.0 when a rule instructs the opposite; in that case set
`"contradicted": true` and cite that rule.

## Instructions

1. **Judge the instruction, not the topic.** Shared vocabulary is not coverage,
   and different vocabulary is not absence. Ask what each text makes the reader
   DO, and whether those actions coincide.
2. **Generalising is not by itself a deduction.** The rulebook is written to
   apply broadly. Score 1.0 if following the general rule still produces the
   specific behaviour. Score 0.5 only when the specific requirement is lost.
3. **Length is not coverage.** A long rule that never gets to the instruction
   is 0.0.
4. **Cite or score zero.** Every score above 0.0 must cite at least one bullet id
   that appears in the rulebook below. If you cannot cite one, the score is 0.0.
5. **Cite at most 3 bullets**, the ones that actually carry the instruction.
6. Fill the fields in the order given: gather evidence, then reason, then commit
   to a score.

## Output format

Return ONLY valid JSON, with the fields in this order:

```json
{
  "evidence": [
    {"bullet_id": "exact id from the rulebook, e.g. gui-00012",
     "quote": "the fragment of that rule that does the work"}
  ],
  "analysis": "1-3 sentences: what the instruction requires, what the cited rules require, and where they diverge",
  "contradicted": false,
  "score": 1.0
}
```

## The rulebook (complete)

<THE COMPLETE LEARNED PLAYBOOK>

## The instruction to assess

- **instruction:** <ONE ATOMIC EXPERT SKILL>
\end{lstlisting}


\section{Example Conversation Replay}
\label{sec:appendix_replay}

This appendix replays one complete episode of the offline benchmark
(\S\ref{ssec:offline}), so that the abstract quantities of
\S\ref{sec:results}---decision points, cited rule identifiers, judge states,
verdicts, and the mutual-silence fast path---can be read against a concrete
conversation. The episode is a negotiation task ($\tau_{\text{neg}}$)
facilitated live by the expert prompt $\prompt^*_{\text{neg}}$. It was chosen
because it is short (twelve messages, six decision points) and because its six
decisions exhibit every verdict type the benchmark produces.
The replay is off-policy in exactly the sense of the structural caveat in
\S\ref{ssec:offline}: at every decision point the learned playbook is asked
what it \emph{would} do given the expert's prefix, its answer is judged, and
the transcript then continues with what the expert \emph{actually} did.

\paragraph{Transcript.}
Table~\ref{tab:app_replay_transcript} gives the full chat stage. The clock is
elapsed time since the scripted welcome. Participants are shown by their
negotiation role; the platform's system messages name the departing
participants by their anonymous animal alias, and in this cohort Party~A's
alias was Monkey and Party~B's was Dog. The three facilitator messages after the
welcome (messages 5, 7, and 10) are the expert's logged interventions and are
therefore also the reference actions $a_i^{\text{ref}}$ at the corresponding
decision points.

\begin{table}[htbp]
\centering
\footnotesize
\setlength{\tabcolsep}{4pt}
\begin{tabularx}{\linewidth}{@{}rllL@{}}
\toprule
\textbf{\#} & \textbf{Clock} & \textbf{Sender} & \textbf{Message (verbatim)} \\
\midrule
1  & 00:00 & Facilitator (scripted) & Welcome everyone! Over the next 5--10 minutes you'll work together to form an alliance and negotiate how to divide the available funds. If you can all agree on a split you'll walk away with your share as a bonus. Let's get started! \\
2  & 00:16 & Party C & I'm fine spliting it three ways \\
3  & 00:37 & Party B & Yeah I'm also fine with doing that \\
4  & 00:49 & Party A & My preference is a three-way split as well \\
5  & 00:59 & Facilitator (expert) & Since you've agreed on a three-way split of the \$10 please keep in mind that you can only use whole dollar amounts. This means you cannot split it perfectly equally; one person will need to receive \$4 while the others receive \$3. How would you like to distribute those amounts? \\
6  & 01:56 & Party A & I am volunteering at the beginning for \$2 and I would be happy to see you two each get \$4 \\
7  & 02:04 & Facilitator (expert) & Party A has proposed a split where Party A receives \$2 Party B receives \$4 and Party C receives \$4. This totals exactly \$10. Party B and Party C do you agree to these specific amounts? \\
8  & 02:22 & Party C & That works for me \\
9  & 02:39 & Party B & for me as well \\
10 & 02:49 & Facilitator (expert) & Excellent. You have reached a consensus on the \$10 split: Party A receives \$2 Party B receives \$4 and Party C receives \$4. Since everyone is in agreement you can move to the next stage by clicking the button in the bottom right once it becomes available at the 5-minute mark. \\
11 & 05:05 & System & Monkey has left the chat. \\
12 & 05:05 & System & Dog 2 has left the chat. \\
\bottomrule
\end{tabularx}
\caption{The replayed negotiation episode. Messages 5, 7, and 10 are the
expert facilitator's logged interventions; the group closed a valid
\$2/\$4/\$4 three-way split of the \$10 pot.}
\label{tab:app_replay_transcript}
\end{table}

\end{document}